\documentclass[twoside,twocolumn,9pt]{article}
\usepackage{extsizes}
\usepackage[super,sort&compress,comma]{natbib} 
\usepackage[version=3]{mhchem}
\usepackage[left=1.5cm, right=1.5cm, top=1.785cm, bottom=2.0cm]{geometry}
\usepackage{balance}
\usepackage{mathptmx}
\usepackage{amssymb}
\usepackage{sectsty}
\usepackage{graphicx} 
\usepackage{lastpage}
\usepackage[format=plain,justification=justified,singlelinecheck=false,font={stretch=1.125,small,sf},labelfont=bf,labelsep=space]{caption}
\usepackage{float}
\usepackage{fancyhdr}
\usepackage{fnpos}
\usepackage[english]{babel}
\addto{\captionsenglish}{%
  
}
\usepackage{array}
\usepackage{droidsans}
\usepackage{charter}
\usepackage[T1]{fontenc}
\usepackage[usenames,dvipsnames]{xcolor}
\usepackage{setspace}
\usepackage[compact]{titlesec}
\usepackage{hyperref}
\usepackage{mathrsfs}
\usepackage{amssymb}
\usepackage{chemarrow}

\usepackage{epstopdf}

\definecolor{cream}{RGB}{222,217,201}

\begin{document}

\pagestyle{fancy}
\thispagestyle{plain}
\fancypagestyle{plain}{
\renewcommand{\headrulewidth}{0pt}
}

\makeFNbottom
\makeatletter
\renewcommand\LARGE{\@setfontsize\LARGE{15pt}{17}}
\renewcommand\Large{\@setfontsize\Large{12pt}{14}}
\renewcommand\large{\@setfontsize\large{10pt}{12}}
\renewcommand\footnotesize{\@setfontsize\footnotesize{7pt}{10}}
\makeatother

\renewcommand{\thefootnote}{\fnsymbol{footnote}}
\renewcommand\footnoterule{\vspace*{1pt}%
\color{cream}\hrule width 3.5in height 0.4pt \color{black}\vspace*{5pt}} 
\setcounter{secnumdepth}{5}

\makeatletter 
\renewcommand\@biblabel[1]{#1}            
\renewcommand\@makefntext[1]%
{\noindent\makebox[0pt][r]{\@thefnmark\,}#1}
\makeatother 
\renewcommand{\figurename}{\small{Fig.}~}
\sectionfont{\sffamily\Large}
\subsectionfont{\normalsize}
\subsubsectionfont{\bf}
\setstretch{1.125} 
\setlength{\skip\footins}{0.8cm}
\setlength{\footnotesep}{0.25cm}
\setlength{\jot}{10pt}
\titlespacing*{\section}{0pt}{4pt}{4pt}
\titlespacing*{\subsection}{0pt}{15pt}{1pt}
\renewcommand{\vec}[1]{\mathbf{#1}}
\newcommand{\g}[1]{\mathfrak{#1}}

\fancyfoot{}
\fancyfoot[LO,RE]{\vspace{-7.1pt}\includegraphics[height=9pt]{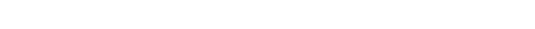}}
\fancyfoot[CO]{\vspace{-7.1pt}\hspace{13.2cm}\includegraphics{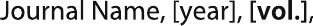}}
\fancyfoot[CE]{\vspace{-7.2pt}\hspace{-14.2cm}\includegraphics{head_foot/RF}}
\fancyfoot[RO]{\footnotesize{\sffamily{1--\pageref{LastPage} ~\textbar  \hspace{2pt}\thepage}}}
\fancyfoot[LE]{\footnotesize{\sffamily{\thepage~\textbar\hspace{3.45cm} 1--\pageref{LastPage}}}}
\fancyhead{}
\renewcommand{\headrulewidth}{0pt} 
\renewcommand{\footrulewidth}{0pt}
\setlength{\arrayrulewidth}{1pt}
\setlength{\columnsep}{6.5mm}
\setlength\bibsep{1pt}

\makeatletter 
\newlength{\figrulesep} 
\setlength{\figrulesep}{0.5\textfloatsep} 

\newcommand{\topfigrule}{\vspace*{-1pt}%
\noindent{\color{cream}\rule[-\figrulesep]{\columnwidth}{1.5pt}} }

\newcommand{\botfigrule}{\vspace*{-2pt}%
\noindent{\color{cream}\rule[\figrulesep]{\columnwidth}{1.5pt}} }

\newcommand{\dblfigrule}{\vspace*{-1pt}%
\noindent{\color{cream}\rule[-\figrulesep]{\textwidth}{1.5pt}} }

\makeatother

\twocolumn[
  \begin{@twocolumnfalse}
{
}\par
\vspace{1em}
\sffamily
\begin{tabular}{m{4.5cm} p{13.5cm} }

\includegraphics{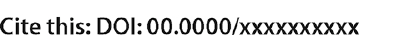} & \noindent\LARGE{\textbf{Icospherical Chemical Objects (ICOs) allow for chemical data augmentation and maintain rotational, translation and permutation invariance.$^\dag$}} \\
\vspace{0.3cm} & \vspace{0.3cm} \\

 & \noindent\large{Ella M. Gale$^{\ast}$\textit{$^{a}$} } \\

\includegraphics{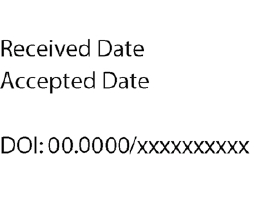} & \noindent\normalsize{Dataset augmentation is a common way to deal with small datasets; Chemistry datasets are often small. Spherical convolutional neural networks (SphNNs) and Icosahedral neural networks (IcoNNs) are a type of geometric machine learning algorithm that maintains rotational symmetry.  Molecular structure has rotational invariance and is inherently 3-D, and thus we need 3-D encoding methods to input molecular structure into machine learning. In this paper I present Icospherical Chemical Objects (ICOs) that enable the encoding of 3-D data in a rotationally invariant way which works with spherical or icosahedral neural networks and allows for dataset augmentation. I demonstrate the ICO featurisation method on the following tasks: predicting general molecular properties, predicting solubility of drug like molecules and the protein binding problem and find that ICO and SphNNs perform well on all problems. } \\

\end{tabular}

 \end{@twocolumnfalse} \vspace{0.6cm}

  ]

\renewcommand*\rmdefault{bch}\normalfont\upshape
\rmfamily
\section*{}
\vspace{-1cm}


\footnotetext{\textit{$^{a}$~Address, Address, Town, Country. Fax: XX XXXX XXXX; Tel: XX XXXX XXXX; E-mail: xxxx@aaa.bbb.ccc}}
\footnotetext{\textit{$^{b}$~Address, Address, Town, Country. }}

\footnotetext{\dag~Electronic Supplementary Information (ESI) available: [details of any supplementary information available should be included here]. See DOI: 00.0000/00000000.}

\footnotetext{\ddag~Additional footnotes to the title and authors can be included \textit{e.g.}\ `Present address:' or `These authors contributed equally to this work' as above using the symbols: \ddag, \textsection, and \P. Please place the appropriate symbol next to the author's name and include a \texttt{\textbackslash footnotetext} entry in the the correct place in the list.}










\section{Introduction}



Machine learning has been applied to many tasks with great success in recent years. After Alexander Krizhevsky won the ImageNet challenge\cite{ImageNet} with AlexNet,\cite{AlexNet} a convolutional neural network (CNN), the field of neural networks took off. AlexNet was a prototypal system. Trained with a large variety of labelled image data taken from the internet, utilising the novel hardware of GPUs, and built of many layers including five convolutional layers, pooling layers and 3 feed-forward layers. After AlexNet's success, deeper CNNs were built with many more layers, for example Inception\cite{47} with over 100 layers, and more different component layers were invented. These models were so successful that the opinion that `image recognition is a solved problem' could be heard at conferences. Some machine learning researchers then started to look for the next big challenge and alighted on chemistry, helpfully, just as chemists were starting to take notice of what machine learning might do for their field. 

But chemistry has different challenges to the image recognition task. A simple (and perhaps tongue-in-cheek) description of the modern machine learning method is to throw vast amounts of data at an algorithm in the hope that the algorithm will be complex enough to learn what is important from the data to solve the problem (The opposite approach to this is feature engineering where an expert assembles what they believe is important in the data, and this was often combined with symbolic A.I.). This is why NNs took 40 years to go from an interesting idea to a crucial technology: we needed to wait for powerful computers, large data storage and large amounts of easily accessible data. However, chemistry datasets are tiny. They are tiny compared to the size of chemical space, for example, it has been suggested that there are around 10$^{60}$ possibly pharmacologically active molecules, mankind has done very well to synthesise 10$^{14}$ but that is still a tiny fraction. Secondly, chemistry datasets are tiny compared with the size of datasets used in image recognition, for example, a simple dataset used in tutorials is MNIST which is only 60,000 images for 10 image classes (the handwritten didgets from 0-9), ImageNet is 1.3 millions images in 1000 classes. The size of the benchmark datasets in MoleculeNet\cite{wu2018moleculenet} range from 642 examples to around 400,000 examples. For example, the 2020 version of the PDBBind dataset has only 23,496 structures in it (both protein binding pockets and ligands) and from this tiny dataset we are supposed to solve a problem as complex as predicting the binding energy!

In my recent paper,\cite{GaleTopolJournal} I demonstrated how a very simple and tiny neural network could perform as well on a task as very complex and large NNs if the inputs (features) are contain just the information necessary to solve the problem. In that work, I used the tools of persistent homology to encode shape properties and tested it on the protein binding problem. In this work, I use the tools of symmetry to attempt to deal with the problem of small datasets and find a new way to encode 3-D structure. 

\subsection{Symmetry} 

\begin{table*}[htp]
    \centering
    \begin{tabular}{|c|c|c|}
    \hline 
        Architecture & Domain, $\Omega$ & Symmetry Group $\g{G}$\\
        \hline
        Convolutional neural networks (CNN) & Grid & Translation \\
        Spherical CNN (SphNN) & Sphere / SO(3) & Rotation \\
        Graph neural networks (GNN) & Graph & Permutation $\sum_n$ \\
        \hline
    \end{tabular}
    \caption{Selected model architectures useful to chemists and their symmetries. Adapted from Geometric Machine Learning~\cite{bronstein2021geometric}}
    \label{tab:model_architectures}
\end{table*}

This discussion frames machine learning in terms of \textit{geometric machine learning}, a novel synthesizing approach to the theoretical side of the subject, which uses geometrical principles, primarily the study of symmetry and symmetry invariants, to categorize machine learning models and approaches.\cite{bronstein2021geometric} Note, this is different and complementary from my other recent work, topological machine learning, which uses persistent homological principles and the study of topological invariants.

Machine learning can generally be thought of as mapping from an input space, $X$ (signals/data), to an output space $y$ (labels). It is known that NNs work as function approximators, and we assume that there is a function, $y=f(x)$, that can do this mapping. For example, if we wanted a NN that could identify alcohols from structures, the function would be: `is R-OH present, if yes, y=True, if no, y=False'. The functions that could be learned are affected by an inductive bias called the \textit{prior}, and one very important prior is the symmetry of the \textit{domain} of the problem.

The domain, $\Omega$ is the \textit{space} that $X$ is in, for example, for CNNs you would have $X(\Omega_x)$, where the image data $X$ in the domain of a pixel grid, $\Omega_x$. People rarely write the domain in in these equations, but there is always a domain. So, it is more correct to state that the neural network learns an hypothesis function: 

$$
y(\Omega_y) = F(X(\Omega_x))\,,
$$ 

which relates signals to labels and each are defined in their domain. When solving the problem, the data, $X$, and the domain that either the data is embedded in, $\Omega_x$, will both affect which of the many possible $F$ the NN will learn. Essentially, the symmetry of the domain will also affect the solutions. Problems have their own inherent symmetry, of course. Thus, it makes sense to match the symmetry of the problem, the symmetry of the data and the symmetry of the algorithm for best results.

Table~\ref{tab:model_architectures} relates domain, $\Omega$, to the symmetry group, $\g{G}$. A graph neural network (GNN) operates on the domain of a graph, and this has the symmetry of permutation, i.e. it doesn't matter how you renumber the atoms in a molecular structure, the molecule is still the same. A symmetry of an object is a transformation that leaves a certain property of said object unchanged, i.e. invariant. As symmetries can be invertible, and the inverse is also a symmetry, the collections of all symmetries of an object is a \textit{group} (see appendix ~\ref{sec:maths}). Thus, molecular graphs are invariant to permutation. A grid is a more specialised version of a graph, and this is the domain that a CNN operates on, i.e. the input domain is a grid of pixels (or three grids of pixels for colour pictures). CNNs are famously translationally invariant, i.e. a cat is a cat where-ever it is in the frame. This work concentrates on spherical CNNs, and these possess rotational invariance. The spherical NN has the domain of a sphere, $S^2$, but convolutions must be applied onto the domain $SO(3)$ which is the space of 3D rotations of a sphere (for more details see appendix\ref{sec:maths}). As molecules have rotational symmetry, SphNNs (and IcoNNs) are a natural choice for problems involving molecular structure.\footnote{For small molecules anyway. A CNN would be a more natural choice for problems involving surfaces.}


\begin{figure*}[htp]
    \centering
    \begin{tabular}{ccccc}

    \includegraphics[width=3cm]{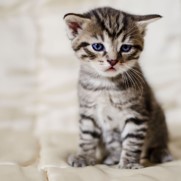} &
    \includegraphics[width=3cm]{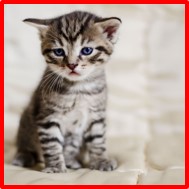} &
    \includegraphics[width=3cm]{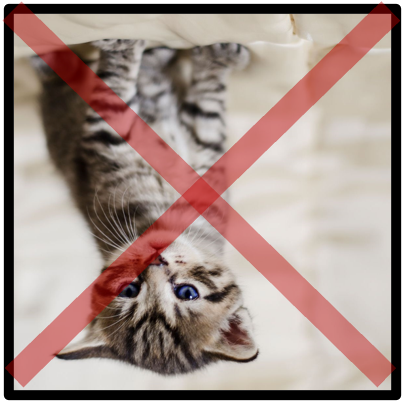} &
    \includegraphics[width=3cm]{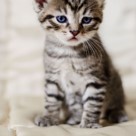} &
    \includegraphics[width=3cm]{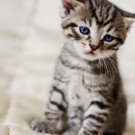}\\
    \includegraphics[width=3cm]{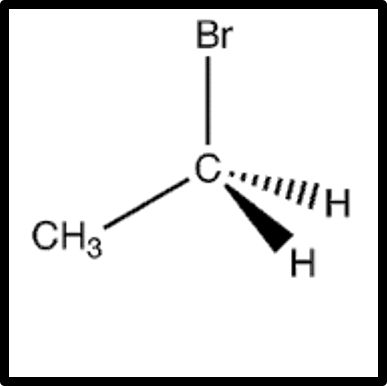}&
    \includegraphics[width=3cm]{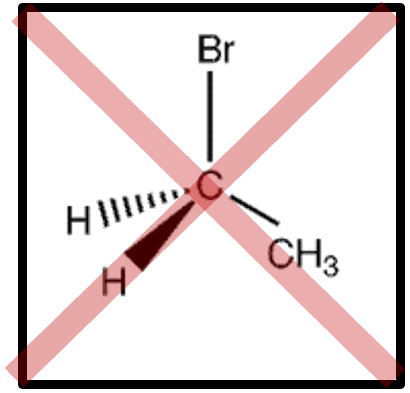}&
    \includegraphics[width=3cm]{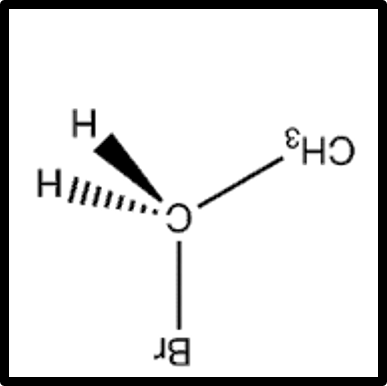}&
    \includegraphics[width=3cm]{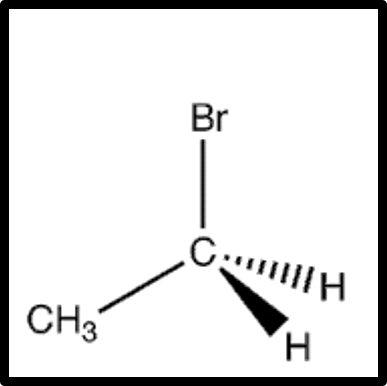}&

    \end{tabular}
    \caption{Using symmetry properties for augmentation. Top row, image augmentations: original image, vertical reflection, horizontal reflection, zoom and crop, rotate a small amount, zoom and crop. Bottom row, molecular augmentations: reflection, rotation, translation.}
    \label{fig:augmentation}
\end{figure*}

\subsection{Spherical and icosahedral neural networks.}

Convolutional layers work by sliding a small set of square filters over the image and from that learning useful features from which the NN can learn how to identify the content of an image.\footnote{Actually, there are many filters which look at the whole image at once and share weights, but the sliding filter schema is a simpler explanation for how translational invariance is implemented.} Hoogeboom et al\cite{hoogeboom2018hexaconv} expanded the idea of convolutional levels to work with hexagonal filters and these filters are the basis of icosahedral CNNs\cite{cohen2019gauge} in which 3D object data is projected onto an icosahedron, which is then unfolded into nets with triangular pixels, which, to make these compatible with standard CNNs (and thus faster to train) are then converted to square pixel charts for input into a CNN. Spherical neural networks\cite{cohen2018spherical, lee2019spherephd} dispense with the conversion to square pixels and work with triangular pixels to input icospherical nets. A connection matrix recording how the triangular pixels are connected at the edge of the nets is input with the data so the convolutional filters can move over the icosahedral net boundaries, effectively joining the nets back into a sphere inside the NN. These NNs preserve symmetries that are useful for chemistry. The hexagonal convolutions are rotationally invariant, and the gauge equivalent CNNs preserve gauge symmetries. These CNNs have found uses in areas where these types of symmetries are important, such as interpreting data from aerial photographs\cite{hoogeboom2018hexaconv} (as these images have no `up' or `down',  omnicameras\cite{lee2019spherephd} (used in self driving cars and robotics), 3D object recognition\cite{cohen2018spherical} and processing astronomical and global geographical and weather data\cite{cohen2019gauge} (gauge symmetry preserves the path independence of vectors that are moved over a curved surfaces). The preservation of rotational symmetry is necessary for molecular input, if we can convert molecular structure to spherical inputs we can make use of these spherical and icosahedral CNNs. 

\subsection{Augmentation}

For large NNs to learn effectively, lots of data is needed. If you cannot get more data, you can augment your dataset by deriving new datapoints from the data you have. In CNNs, you can create new datapoints that have a different pixel space $X(\Omega_x)$, by reflecting, cropping or slightly rotating the source image, see the cat pictures in figure~\ref{fig:augmentation}, the pixels are different, but the image still contains a cat. We can apply this idea to molecular structure, but we must attend to the symmetries of molecules. For example, we do not want to reflect the structure as this will be a different enantomer in chiral molecules, and for most molecular machine learning tasks we would want to preserve chirality. Unlike the CNN version, we can rotate the molecule freely and it is still the same molecule.\footnote{The reader might have realised that there is no reason as strong as chirality preservation to prevent us from reflecting the cat along a horizontal axis, after all an upside down picture of a cat is still a cat. Generally, however, images do have a correct way up, and the CNN researchers choose to try to keep that property when training. As an upside down picture of a cat is still a valid picture of a cat, but it isn't as useful an example as one rotated by a small amount.} Therefore, we if we use spherical or icospherical neural networks, we can augment our molecular datasets whilst preserving the important symmetries of the data.

\subsection{Symbols}
 Unlike the feature engineering approaches mentioned earlier, here we are not using symbolic A.I. with them. This is largely a matter of choice. The large CNNs worked well without symbols, so it is a good idea to keep the power of the nonsymbolic, statistical approach at the level of ML algorithm. Furthermore, chemistry is full of symbols that serve as a useful shorthand to human chemists (like bonds, atom labels, charges, valence etc), some of which are ill-defined as simple rules to define them are often broken by a few counter-examples. Instead, we choose from nature's atom labels, i.e. the atomic mass and number for each element, and hope that from this data combined with the geometrical arrangement of the atoms the SphNN is able to learn about chemical concepts like bonds, the difference between sp$^2$ and sp$^3$ hybridised carbons and so on. If this technique did not work, or if the reader would prefer atom labels the icospherical maps can be prepared with labels instead. Note, that by not having atom labels, we also implicity do not have atom numbers (as in, node numbers in the molecular graph) so this technique is permutation invariant as well as rotationally and translationally invariant. 

 \subsection{Featurisation and dimensionality}

It is not obvious to know what is the best featursiation for a molecule, and more to the point, it will be different for different tasks. I posit that the 3-D structure of a molecule is essential for the tasks considered here. Some common featurisations are SMILES strings (and related ideas like InChi, SMARTS, SMIRKS etc) that encode the molecular formula. These are often chosen as they look information dense and simple, and their similarity to language (SMILEs strings do possess a simple grammar that encodes chemical rules) has prompted people to apply language models to them. However, chemical reactions and formulae are not a langauge.\cite{balllang, gordinlang}

If the 3-D structure of a molecule is important to the task, then by using SMILES strings we are asking the NN to learn about both 3-D space and the behaviour of molecules in that 3-D space\footnote{By which I mean molecular geometry, for example the tendency of sp$^3$ carbons to have tetrahedrally arranged bonds or the tendency for C-H bonds to be shorted the C-C bonds.} from 1-D space. The way a student human chemist generally works out a structure from a formula is as follows, after reading the 1-D symbolic formula (e.g. C$_6$H$_6$) they would first draw or think about the 2-D molecular graph (i.e. it is a hexagon as this molecule is cyclohexane) and to do this requires chemical knowledge in that they need to know what the symbols `C' and `H' refer to in order to figure out the connectivity. Then, they need to embed that 2-D molecular graph in a 3-D space and to do that they apply both their knowledge of both how 3-D space works and the 3-D structure carbon atoms (i.e. that the bonds will be in a tetrahedral arrangement around carbon) to start drawing a wavy, bent hexagon which is how we tend to draw cyclohexane. I do not believe that this student would find it easy to then realise that there are conformers (boat, chair and twisted boat etc) without playing with a ball and stick models in a 3-D space (i.e. the real world space or a virtual reality space). This is a rather involved process that has involved embedding from 1-D to 2-D, then 2-D to 3-D with a large amount of chemical knowledge required along with an understanding of how 2-D and 3-D spaces work. This is a difficult task for a NN that has never worked in or with 2-D and 3-D spaces. If we think about human development, both of a human baby and the human race, we can immediately that 3-D space is the more natural space to use to define molecular structure: a human child learns to see and understand 3-D space at around 6 months, at 18 months they start drawing objects around them, learning how to embed a representation of a 3-D object on a 2-D plane, and at 3-4 years old they start mastering the 1-D highly symbolic technology of reading and writing; we can assume that humans have seen and understood 3-D space as long as there were humans, the first cave paintings were 30,000 BCE and the first language appeared around 3400 BCE. Taken together this suggests that to train a A.I. chemist, it would be much easier to give them input from a lowest stage of development, i.e. 3-D shape.\footnote{And to those concerned about artificial general intelligence, AGI, or efficiency in training, making the problem simpler means we can use simpler models and faster, cheaper training.} 

However, the task of inputting 3-D shape into a computer is a difficult problem. The obvious idea of inputting Cartesian coordinates had the problem that a slight shift in atom labelling or the position of the object gives a completely different input: it is not rotationally translationally or permutation invariant. Voxelising the 3-D space has been tried, and this works, but it suffers as the input is sparse (and NNs find it very hard to learn from sparse data) and because NNs need all inputs to be the same size, we must set the maximum size at the start, limiting the range of molecules the model can be shown and futher increasing the sparseness of the input data (a lot more empty space for small molecules). Other methods include starting with 2-D data and doing the embedding for the NN, fingeprints, holograms and rdkit-based molecular properties are examples of this. The Coulomb matrix approach uses 3D data and symmetry functions (doing part of the calculation for the NN as part of the featursation) to create a 2-D input (a matrix) or a 1-D input (the eigenvalues of the matrix). Featurisation approaches are summarised in table~\ref{tab:featurizations}. The ICO approach embeds 3-D structure into a 3-D space\footnote{A reader of my other paper may argue with me here, as the surface of the sphere is a 2-D manifold in topology, but as the spherical NN keeps the 3-D structure of the input and as the convolutions form an $SO(3)$ symmetry which is 3-D, then perhaps we can claim it is geometrically 3-D. The icospherical NNs that unfold the icospheres to 2-D icospherical maps, and then use 2-D convolutaions on them are obviously embedding the problem in a 2-D space, and it is this that offers the speed up with respect to spherical NNs.} (the sphere), which then uses 3-D convolutations on it that maintain rotational and translational symmetry. It also has the advantage of being scale-free and being less sparse an input type. 

\begin{table*}[htp]
\small
  \caption{\ Common and novel featurizations used in molecular machine learning. NN is a feed-forward NN, RF is random forest, KRR is kernel ridge regression.}
  \label{tab:featurizations}
  \begin{tabular*}{\textwidth}{@{\extracolsep{\fill}}llllll}
    \hline
    Type & Dimensionality  & Dimensionality & Examples & Example\\
    Type & of input & of featurization &  & models\\
    \hline
    Chemical formula & 1-D string & 1-D &  SMILES, SMARTS, SMIRKS & MTR or textNN\\
    Fingerprints & 1-D string & 1-D  & ECFP, MACCS & NN, RF, KRR\\
    physicochemical properties  & 1-D & N/A & Rdkit & Multitask Regression\\
    Coulomb matrix & 3-D & 2-D  & Coulomb matrix & DTNN \\
    Eigenvalues of Coulomb matrix & 3-D & 1-D  & CM-Eig & Multitask Regression\\
    Graph & 2-D & 2-D &  Graph convolutions & Graph convolutional\\
    Graph & 2-D & 2-D &  Weave convolutions & neural network\\
    Grid & 3-D &  1-D & Grid input  &MTR\\
    PHF\cite{GaleTopolJournal}  & 3-D & 1-D  & PHFs & Multitask Regression \\ 
    Icospherical/spherical & 3-D & 3-D  & Icospherical maps (this work) &Spherical NN, Ico-NN\\
    \hline
    \label{tab:featurisation}
  \end{tabular*}
\end{table*}

\subsection{Aqueous solubility}

Aqueous solubility is useful for synthetic chemists, especially those working in the pharmaceutical or agricultural fields as it affects the uptake of biologically active compounds in both living creatures and the environment, as drugs must circulate in aqueous medium (such as blood) assessing solubility early in the drug discovery pipeline saves money by directing development towards higher efficacy compounds. Accurate equilibrium solubility determination is a time-consuming experiment, and it is useful to be able to assess solubility in the absence of a physical sample. Solubility is known to be correlated with some chemical features such as logPoctanol (logP) (the octanol/water partitian coefficient)\cite{delaney2004esol}, and features such as the presence of -OH groups (positive correlation) and aromatic rings (negative correlation)\cite{duvenaud2015convolutional}.

Current limitations in predicting solubility have been suggested to be due to very small (and biased) datasets\cite{lusci2013deep}, experimental error and the question of whether the correct molecular descriptors are being used.\cite{hewitt2009silico} Thus there is a clear need to get the most from the solvation datasets and investigate which features are necessary (it was stated in MoleculeNet\cite{wu2018moleculenet} that conformer/ shape was irrelevant, nonetheless we undertake this investigation).

Delaney's paper applied simple linear regression models using the following parameters: clogPs, Molecular weight (MWT) rotatable bonds (RB), 
aromatic proportion (AP), non-carbon proportions, H-bond donor counts, H-bond acceptor counts and  polar surface areas,
and he found that clogP, MWT, RB and AP made significant contributions to the model and got an $R^2$ of 0.69, with a standard error of 1.01 and average absolute error of 0.75 (log solubility in mols per litre).


\subsection{Protein binding problem}

The protein binding problem is a fascinating problem for those interested in 3-D shape and very useful for drug development and design. The task is to score how well a ligand (a drug, perhaps) binds to a protein binding pocket (drug target) and there has been many interesting machine learning models developed to tackle this.\cite{rezaei2019improving,kundu2018machine,jimenez2018k, ballester2014does, jones2020improved, kwon2020ak, soni2020improving, xie2020multitask, zheng2019onionnet, bao2021deepbspl} The PDBBind dataset consists of many protein binding pockets and ligands, the two structures are minimised together and the task is to calculate the binding energy. Usually, workers used the restricted dataset (3000-4000) complexes as the training dataset and the core (285 complexes) as the test. Here we train on the core dataset to show how much can be gotten from the sort of dataset size a small pharamceutical start-up or academic groups might reasonably have access to.

\section{Methodology}

This section will explain how the ICOs are built and the experiments done on them.

\subsection{Icospherical Chemical Objects (ICOs)}

\subsubsection{Overview} We start with a molecular formula, $m$, and from that create a conformer embedded in 3D space. A molecular structure, $x$, is embedded into icospherical space, $x(\Omega_i)$, by projection, and we derive from rotation/unfolding a set of augmented datapoints, $\{x'(\Omega_i)\}$ which are all different in pixel space, $\Omega_i$, but represent the same object, $x$. As the SphNN is rotationally invariant, we expect it to learn:
$$
y = F(x'(\Omega_i))
$$
with rotational invariance. By presenting many pixel-different, but rotationally equivalent $x$, we aim to make it easier for the SphNN to distinguish signal (what is important about chemical structure) from noise (irrelevances like the rotational orientation). We can also further augment the dataset with small random translation and creating a set of different conformers, $x''$ at the embedding into 3-D space stage, and this will give us a set of conformerally and rotationally different embeddings $\{x''(\Omega_i)\}$, which we hope gives the SphNN enough information to learn about flexibility and shape. The featurisation process is then summarised as: 

\begin{equation}
    m \; \autorightarrow{embedding}{in 3D space} x(\mathbb{R}^3) \;
    \autorightarrow{projection}{to icosphere space} x(\Omega_i)
    \; \autorightarrow{rotation}{} \{x'(\Omega_i)\}    
\end{equation}

\subsubsection{Encoding molecular structure via icosahedral projection}

\begin{figure}[htp]
    \centering
    \includegraphics[width=0.3\textwidth,angle=90]{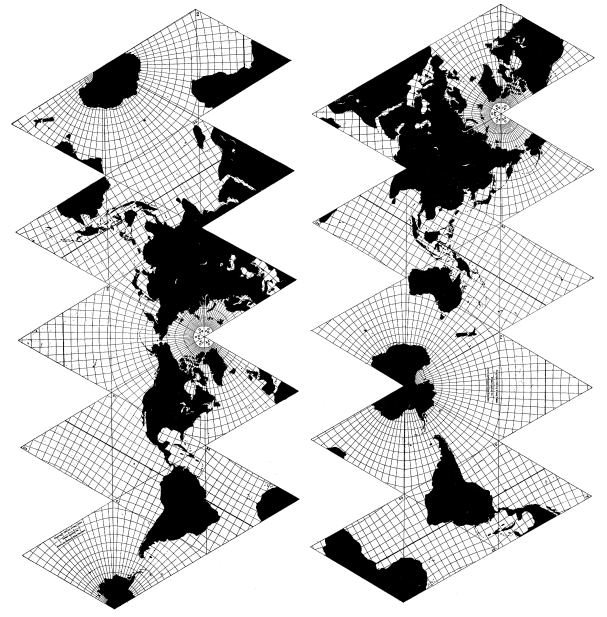}
    \caption{Example of two unfoldings (icosahedron nets) of an icosahedron `globe' of the Earth. There are 60 such nets. Icosahedrons preserve the shape and size of the continents better than projection onto the rectangular plane. Note that if information is split across triangles (like America's landmass in the right hand net) there is also a net where the information is not split (see America on the left-hand net). Taken from \cite{fisher1943world}.} 
    \label{fig:world}
\end{figure}

The icosahedron is the fifth platonic solid and has its own point group: $i_h$. Icosahedrons have 20 equilateral triangular faces. It has been argued that an unfolded the icosahedron is the best 2D representation of a sphere, as the the distortion is the smallest, see figure~\ref{fig:world}, although note that the distortion that does appear is worst at the edges where the triangles would join.\cite{fisher1943world} Icospheres are extensions of the icosahedron where each triangular face is split up into 4 smaller equilateral triangles which are then projected outwards such that each new triangular pixel is the radius of the icosphere distance away from the centre of the sphere (alternatively, you can think of icospheres as tiling a sphere with triangular tiles). This can be repeated to make icospheres of a desired complexity adding in more pixels. Icosphere level 1, ico-1, has 80 pixels; ico-2: 320, ico-3: 1280; ico-4: 5120; and there is a trade-off between the number of pixels (which increases training time and problem difficulty) and the resolution of the input required for the problem (more atoms requires more resolution with the techniques reported herein to get enough shape data to solve the problem).

\subsubsection{Augmenting based on the icosahedron}

\begin{figure}[htp]
    \centering
    \includegraphics[width=0.35\textwidth]{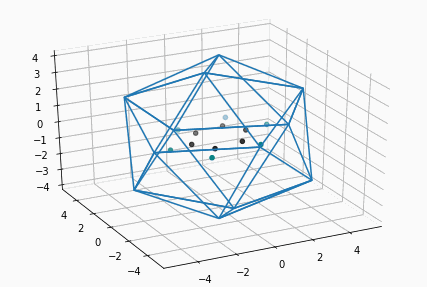}\\
    a. Benzene in icosahedron\\
    \includegraphics[width=0.35\textwidth]{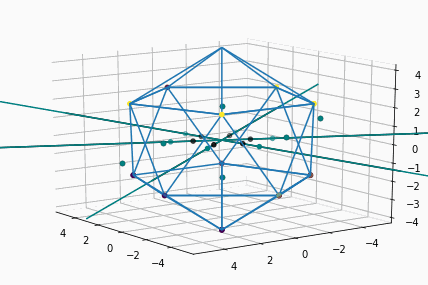}\\
    b. Benzene projected onto icosahedron\\
    \includegraphics[width=0.35\textwidth]{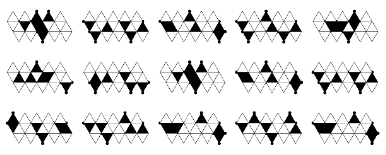}\\
    c. 15 unfolded icosahedral net projections\\
    \caption{Icosahedral projection. Top: Benzene molecule placed at the centre of an icosahedron, note hydrogen atoms are rendered in blue/green. Middle: projecting (ray casting) from the centre point to the icosahedron. Lines are projection directions. Points outside the sphere are the face normals for faces hit by the projection. Bottom: 15 example nets. Note that with this set-up, the hydrogens are not shown as the internal carbons project over them. With rotation this does give the NN valuable information about alignment of bonds. For reactivity and host-guest interactions, projecting so that the outer atoms overwrite the inner might be more appropriate. As benzene has a six-fold rotational symmetry there are repeated patterns and motifs in the nets.}
    \label{fig:map_zB}
\end{figure}

We start by loading a molecular formula (SMILES) or structure (mol2 or pdb). The SMILES strings are turned into 3-D structures using Rdkit,\cite{rdkit} adding the hydrogens\footnote{This is a design choice. For a NN to learn about chemistry, especially reactivity (which is where this work is heading) I think it needs to `see' the hydrogens which are often the whole or part of the groups moving around during a reaction. Trying to infer about hydrogens from just the heavy atom backbones of molecules is a harder problem than being given them, so in the spirit of making the problem easier, we keep the hydrogens in.} and doing a very simple minimisation using the UFF forcefield. The molecular stucture is then placed at the centre of an icosphere, see figure~\ref{fig:map_zB}a. The atom's positions are then projected onto the surface of the icosphere using ray-casting, we can imagine a light source at the centre of the icosphere and then imagine that the atoms are casting shadows onto the icosphere, see figure~\ref{fig:map_zB}b. At this point, we can unfold the icosahedron to an icosahedral net in one of 60 ways and each of these unfoldings is equivalent to a rotation, see figure~\ref{fig:map_zB}c. For icosahdral NNs we would use these unfolded nets as the inputs, for spherial NNs these nets are not unfolded but input as is. In both cases this results in data augmentation, as we now have 60 nets/spheres that differ in pixel space but represent the same object. This helps the NN learn what the actual important features of the input object are, i.e. it is the molecular structure, not the pixel embedding of that structure. 

\subsubsection{Symmetry features retained by the embedding}

\begin{figure*}[htp]
    \centering
    \includegraphics[width=\textwidth]{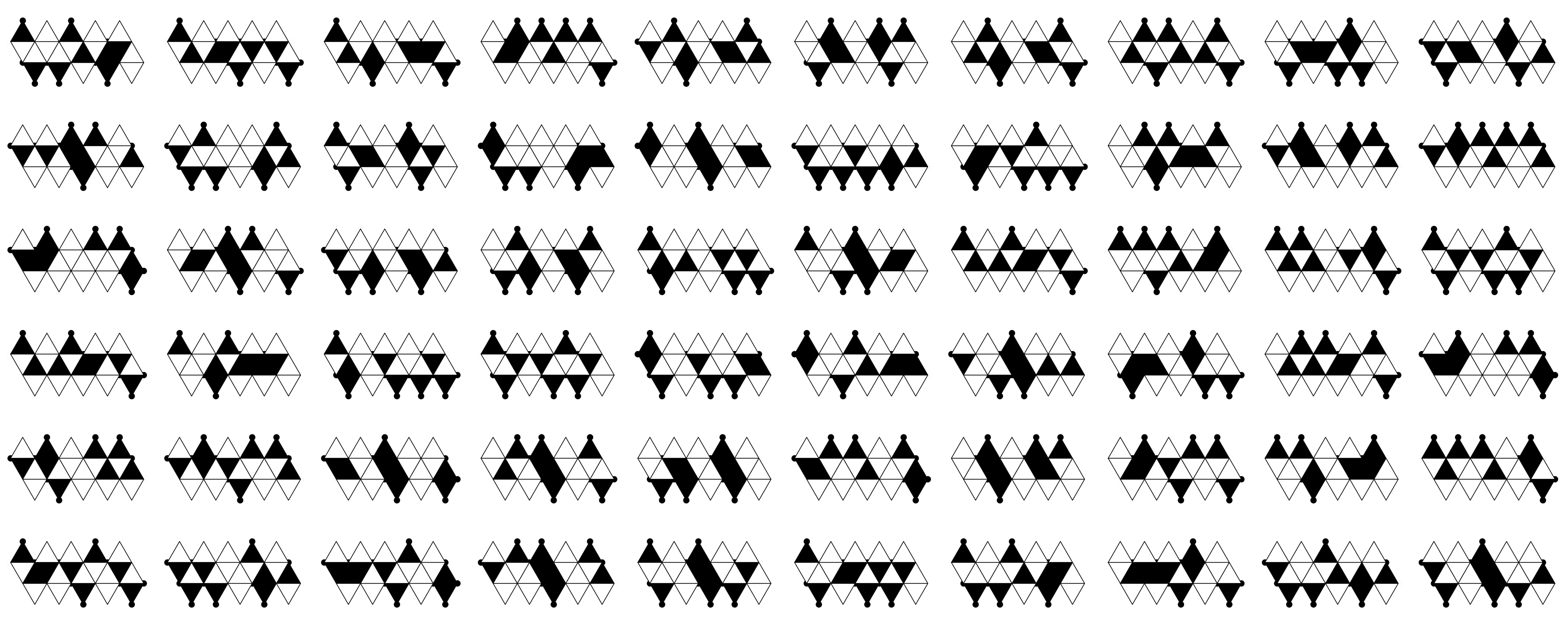}
    \caption{Symmetries interacting: cubane's cuboidal symmetry gives rise to rotational symmetry when projected onto a icosphere. Black pixels are those which the carbon atoms project onto. The ICOs can be unfolded to 2-D (as here) for input into icospherical NNs or if input as icospheres they correspond to a small rotation of the molecule.}
    \label{fig:cubane}
\end{figure*}

\begin{figure*}[htp]
    \centering
    \includegraphics[width=6cm]{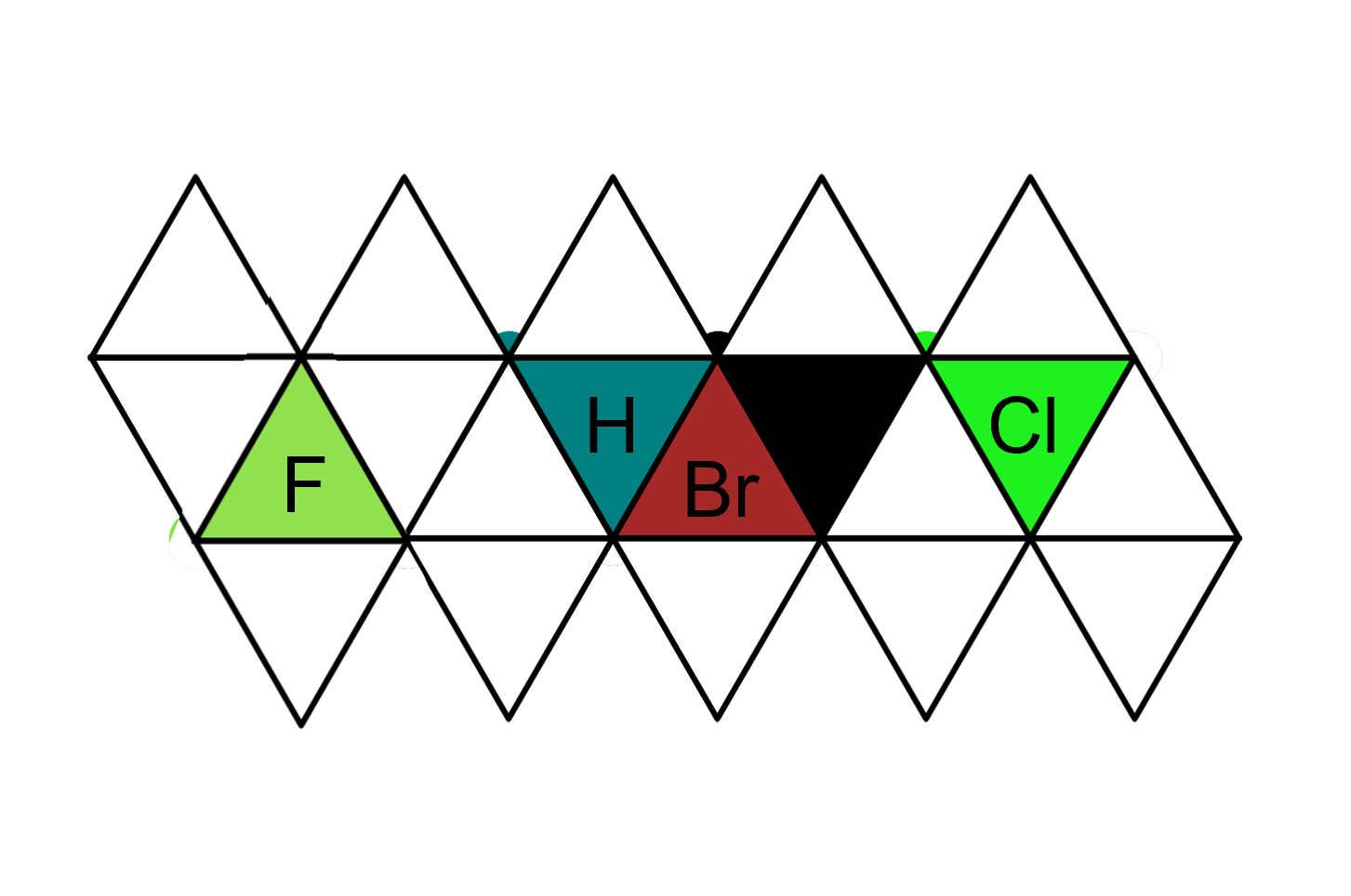}
    \includegraphics[width=6cm]{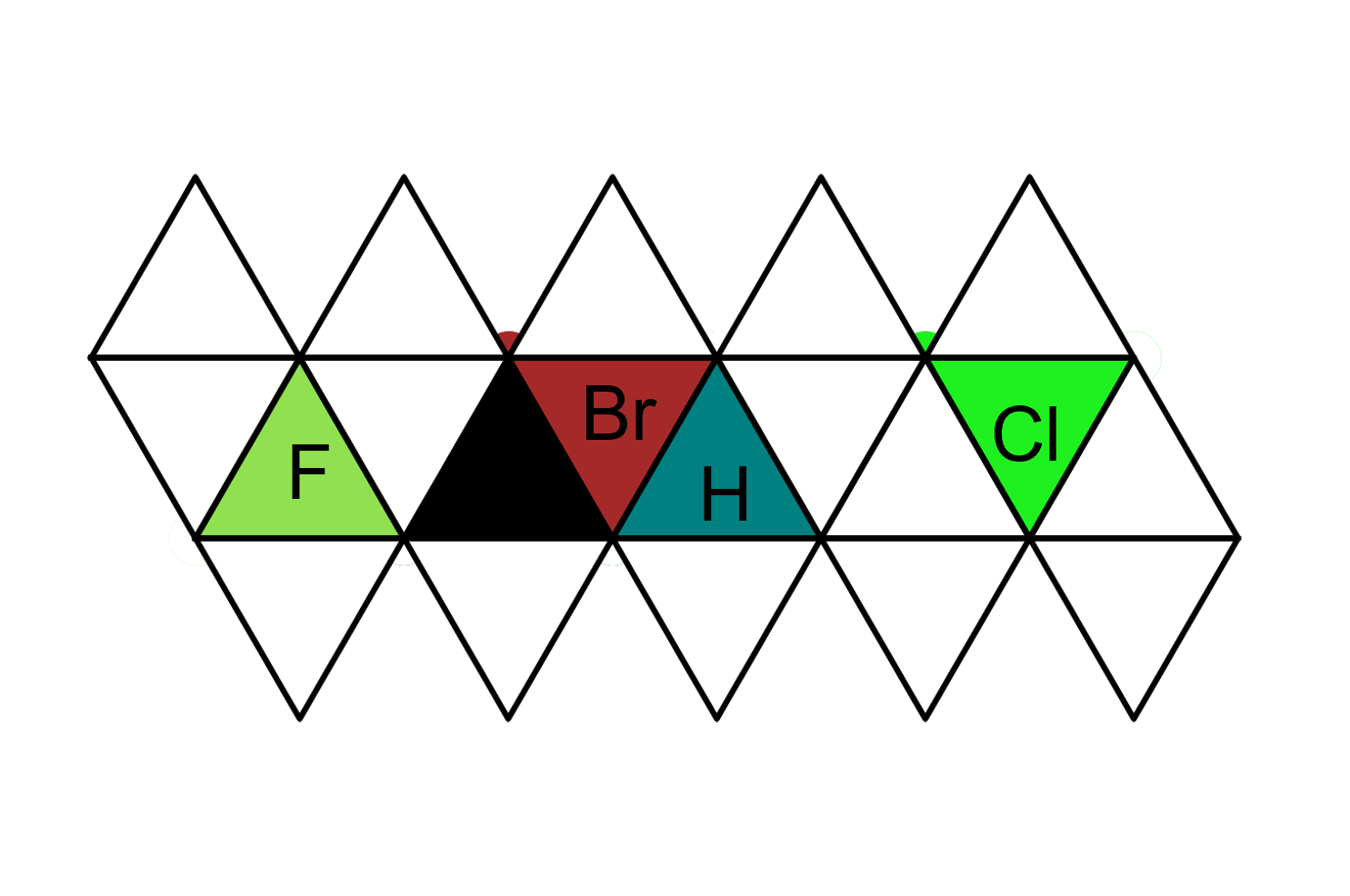}\\
    \includegraphics[width=13cm]{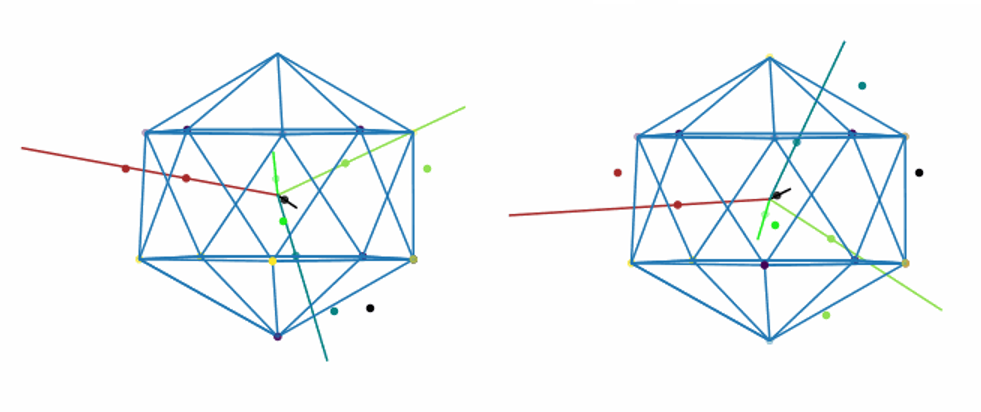}
    \caption{ICOSTAR preserves chirality. Left: R-BrClFmethane, right: S-BrClFmethane
}
    \label{fig:chiral}
\end{figure*}

Figure~\ref{fig:cubane} shows the 60 maps that result from cubane. Each face of the icosahedron is coloured by the highest mass atom that hit it (i.e. carbon). We can see the interaction of the cuboidal symmetry of the input object, $x$, with the icosahedral domain, $\Omega_i$. There are repeated nets, patterns in the net and rotational simularity between some nets which results from the underlying molecules point group. For example, buckminsterfullerene has the same space group as a dodecahedron and an icosphere, so you could align the bucky ball such that all maps look exactly the same due to the symmetry match between signal and domain. Figure~\ref{fig:chiral} demonstrates that chirality is conserved by this embedding. Instead of a refletion, the nets require a 180 degreee turn and relabelling to match. I have also developed a 3-D embedding that is based on topology and which does not preserve chirality,\cite{GaleTopolJournal} so these two embeddings could be combined to get the best from both approaches.

\subsubsection{Icospheres and fine-graining.}

\begin{figure}
    \centering
    \includegraphics[width=0.4\textwidth]{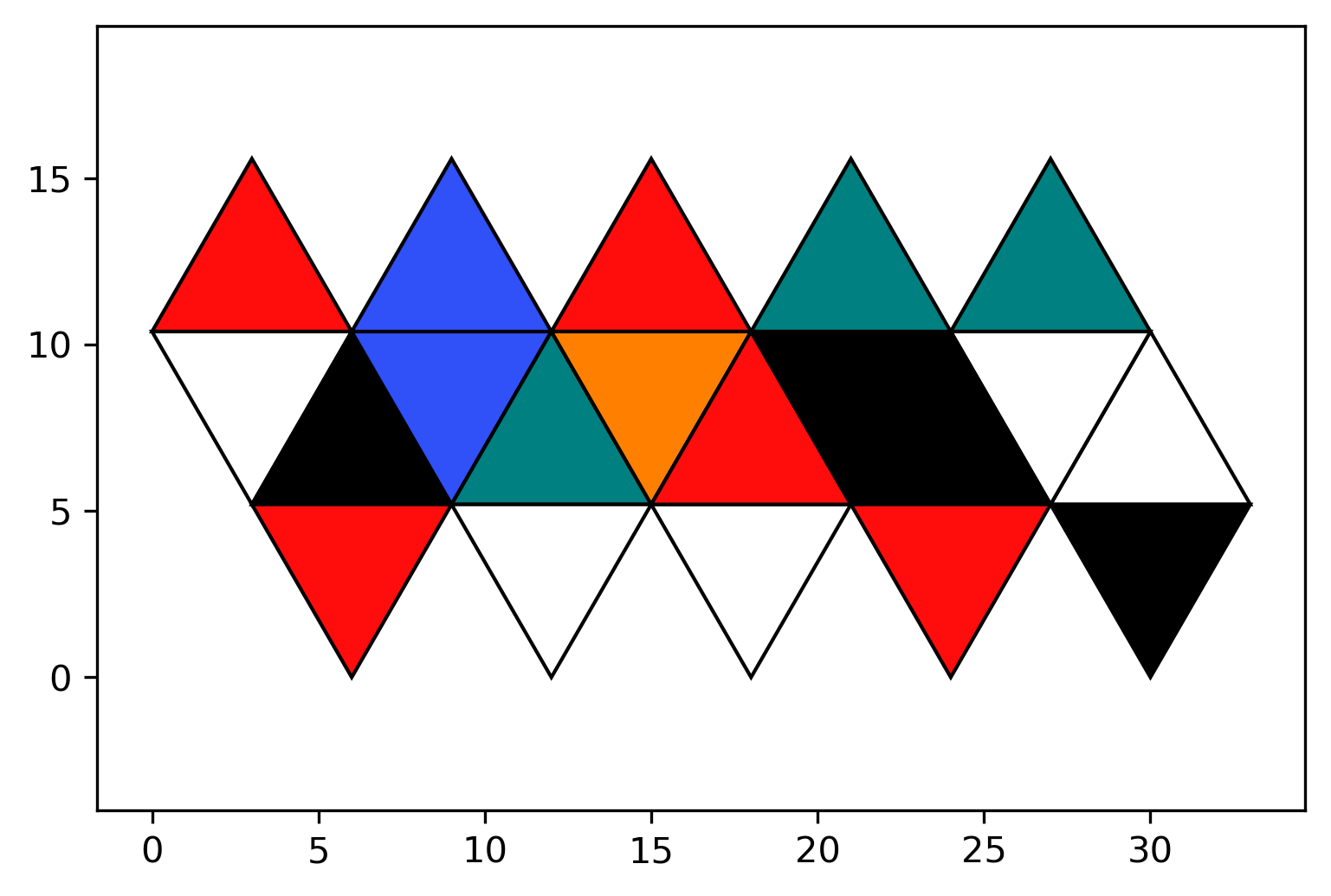} \includegraphics[width=0.4\textwidth]{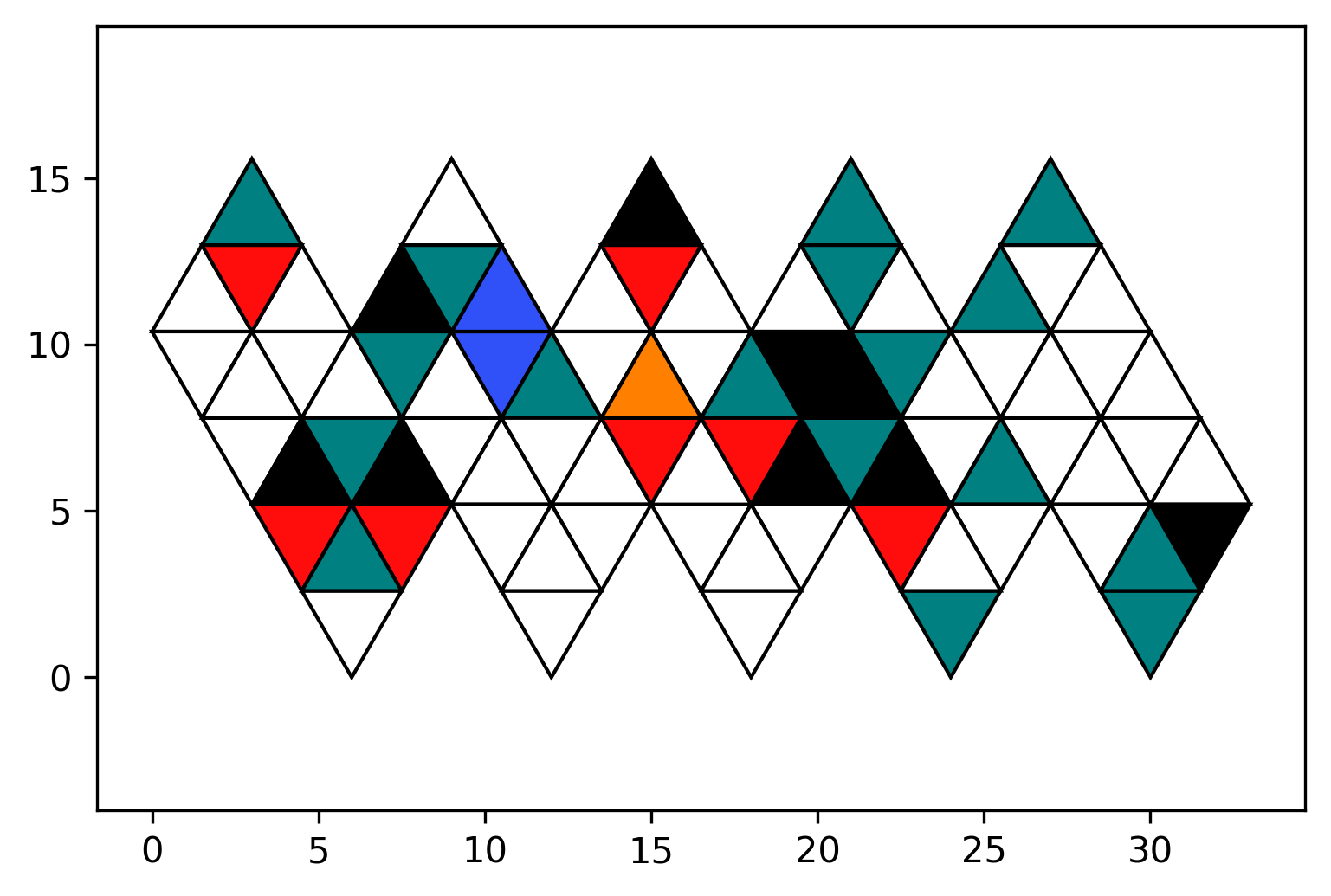}\\
    \includegraphics[width=0.4\textwidth]{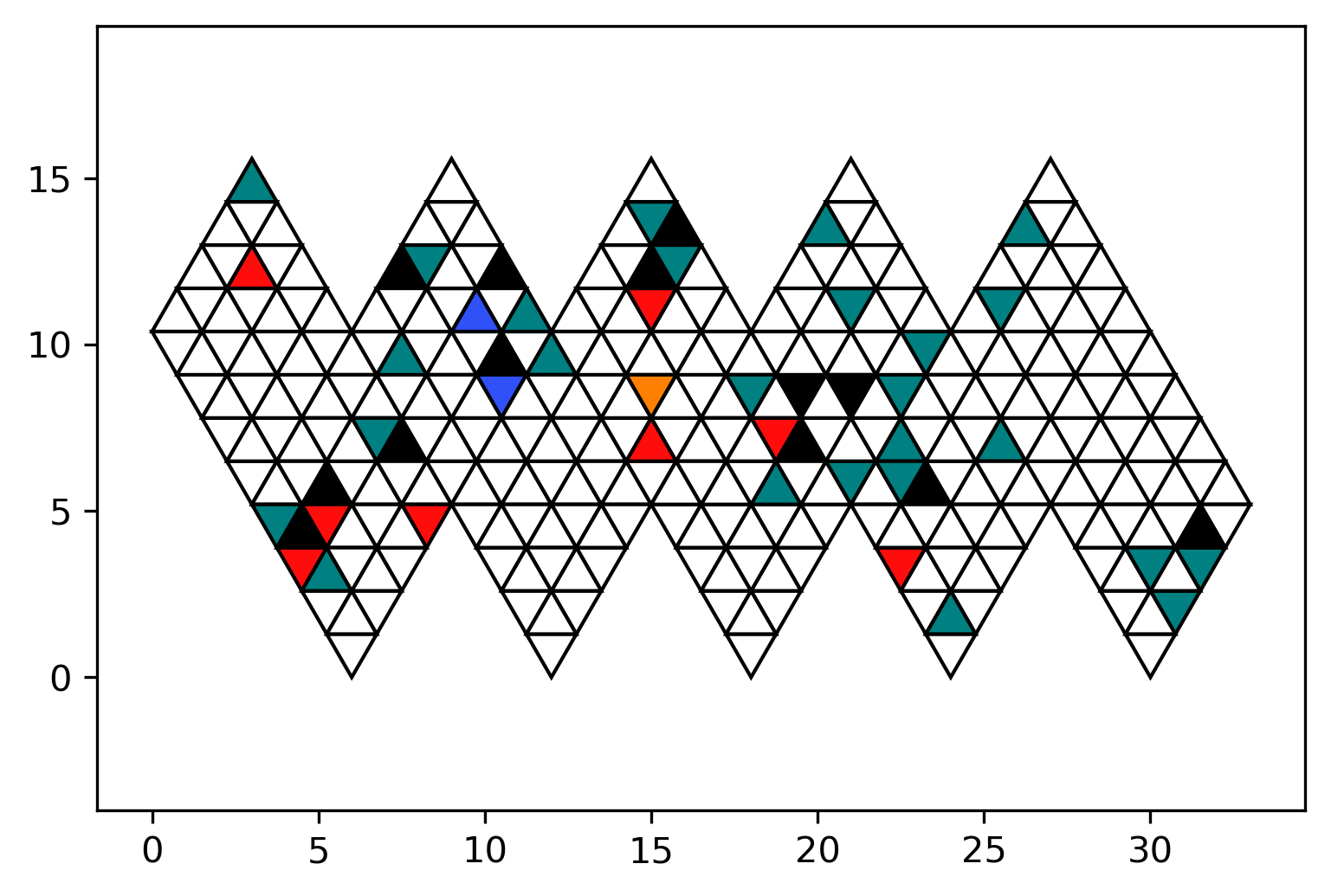}
    \includegraphics[width=0.4\textwidth]{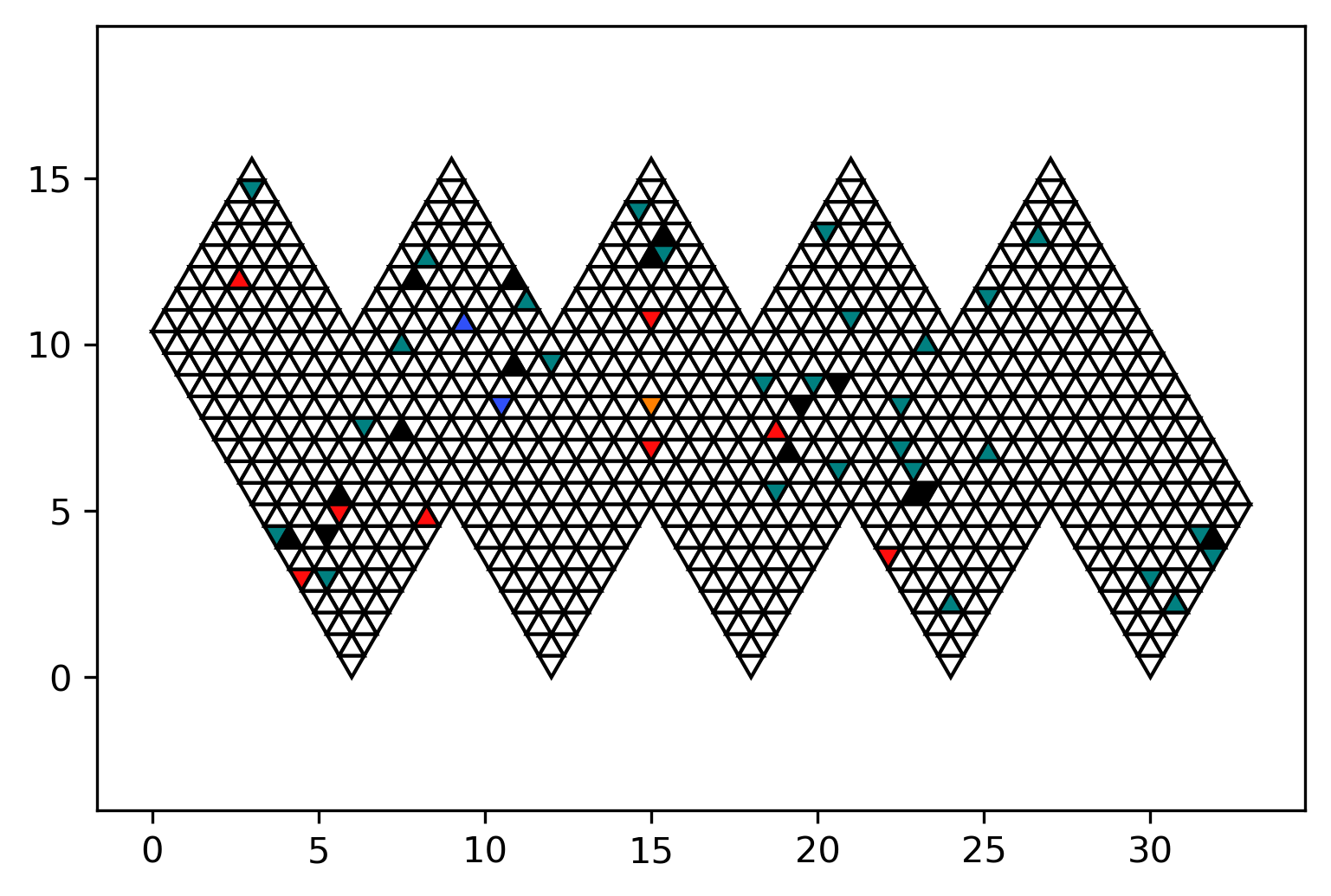}
    
    \caption{Projection onto different levels of icosphere provide different `magnifications' (alternatively, different levels of coarse-graining) of the molecule. The effects of occlusion are show in the bottom right triangle where more atoms are seen with increasing `magnification'. The angles also get more precise with increasing magnification. This example is influenzia medication tamiflu. Hydrogens are coloured teal.}
    \label{fig:tamiflu}
\end{figure}

An icosahdron gives us only 20 pixels, but from an icosahedron you can create icospheres. The faces are subdivided into 4 equilateral triangles and then projected out to lie 1 radius from the centre. This can then be repeated to give you icospheres of different levels, with more equilateral triangular pixels. With these different levels of icosphere we can get more fine-grained representation of the molecule. Figure~\ref{fig:tamiflu} shows the icosaspherical nets for the icosahedron and the first (ico-1), second (ico-2) and third (ico-3) level icospherical embeddings of the same molecular structure. We can see the refinement of detail in the structure. Animated movies of the paths the atoms take across the surface of the icospherical net as the molecule rotates have been produced (see S.I.). The atoms dance their way across the net, with atoms in the same functional group travelling close to each other, and each atom tracing out its own path dependent on it's position and geometry in the molecule. For example, the sulfur atom (orange) performs a swirling, looping dance across the surface. Note that, adding in more pixels gives many more possible unfoldings for the icosphere, for simplicity we chose to keep the original 60 unfoldings.

\subsubsection{Colouring in the pixels}

In the examples thus shown the pixels are coloured according to the largest mass atom whose ray was incident on the face (pixel) of the icosphere. For the machine learning we could input the pictures, but instead we chose to use nature's atomic labels, namely the masses of the atoms. Input images have 3 dimensions, R, G, B (or red, green, blue), there is nothing stopping us from adding more, but for this work these three were the only used. We take the red channel to be the mass of the inner most atom (i.e. that closest to the centre of the icosphere) whose ray as incident on the pixel. G is the mass of the outer most atom (furthest from the centre of the icosphere). And B is the sum of all atoms incident on that pixel. For a single atom, e.g. oxygen, we would have O = [15.999, 15.999, 15,999]. A carbonyl group could be CO=[12.0107, 15.999, 28.0097 ], if the projection was along the bond and carbon was closer to the middle of the molecule. O-H could be [15.999, 1.0078, 17.0068] C-H could be [12.0107, 1.0078, 13.0185] and so on. Obviously, some atoms are only present in the B channel, although for small molecules, as the molecule turns, the hidden atoms will occasionally be incident on the icosphere. In the case of proteins this obviously means that you lose a lot of data, but this is similar to protein manifold techniques. 

It was chosen to use the actual atomic weight and not the integer floor to allow the NN to differentiate between different groups. For example, if we took O as 16, C and 12 and H as 1, a coincident 17 in the blue channel could come from $16 + 1$ (O and H) or $12 + 1 + 1 + 1+ 1+ 1$ (C and 5 hydrogens). The inner and outer atoms in the R and G channels also helps the NN to figure out which atoms must have been coincident. Currently, this is the encoding, although there is room to add more channels, for example, adding in charges, or perhaps taking depth slices through the icosphere, or even re-introducting symbols.

\subsubsection{Further augmentation}

By using the level 0 triangles (those from the original icosahedron) there are 60 possible nets corresponding to starting at one of the three directions of each of the 20 triangular faces. Each net is a different input in pixel space generated from the same structure and thus is dataset augmentation. Each net is equivalent to using the same unfolding method and rotating the molecule by a set amount.
At higher levels of icosphere there are many more possible nets but we only take the 60 defined by the original icosahedron. 

There are three further ways to augment the dataset that are included in Icospherical Chemical Objects:\cite{ICO} 
\begin{enumerate}
    \item Rotate the molecule by a random amount between 0$^{\circ}$ and 5$^{\circ}$ around the 3 Cartesian axes (combined with the unfoldings this covers all the rotational angles of the molecule).
    \item Offset the molecule by an amount
    \item Use a different starting conformer. It was thought that using different conformers would implicitly provide information about the flexibility of the molecule as less flexible molecules would have less and more similar conformers, more flexible molecules would have vastly differing conformers. 
\end{enumerate}

\subsection{Constructing the datasets}

The datasets created will fall into the realm of big data, and thus we expect that the data will be too big to fit into memory and operations may need to be done element-wise. Although SphereNN can deal with numpy arrays, in this work everything is exported everything as hdf5 files.

\subsubsection{1. First pass: mean and max}

It is standard in NN research to normalise the dataset to make it easier for the NN to learn and for that it is essential to know the average (mean) net of the dataset. If normalised, the actual input into the SphNN is not the calculated net, but the difference between the calculated net and the overall average net for the dataset. Three types of normalisation are offered: $L^2$, mean and standardisation.

Normalisation works on previously created datafile and requires 3 passes through the dataset (as it's big data). The process is:
\begin{enumerate}
    \item Compute the element-wise mean and max of the dataset
    \item Compute elementwise standard deviation and mean normalisation (using the previously calculated mean) of the datset
    \item Compute the element-wise standardisation of the dataset (using the previously calculated standard deviation). 
\end{enumerate}

The dataset values for each element, $t$, (i.e. triangular face) of the template net are used to calculate the mean elementwise, i.e.
$$
t_{ab} = \sum_{i=0}^{^{(N-1)}/_r} \sum_r \frac{1}{r} x_i,
$$
where $t_{ab}$ is the triangular face on the $a^{\mathrm{th}}$ row, $b^{\mathrm{th}}$ column, $x_i$ is the value on that face for the $i^{\mathrm{th}}$ entry in the dataset and $N$ is the length of the dataset (namely the number of nets). The inner sum is the sum for each batch of size $r$.

Similarly, want the map of max values in the dataset at each face of the net. The max-net is initialised to 0.1 to avoid divide by zero errors in the later steps, and max found via processing batches of data.

\subsubsection{2. Second pass: mean and L2 normalisation, calculation std}

For face $ab$, the mean normalisation $nm$ is calculated as:
$nm_{ab} = x_i - \hat{x_i}$
the $L^2$ normalisation, $nL^2$ is calculated as:
$$
nL^2_{ab} = \left(\frac{(2*x_i)}{u_{ab}}\right) -1
$$
and the standard deviation is calculated similarly to the mean. 

\subsubsection{Pass 3: standardisation}

Then the standardisation is calculated as:

$ns_{ab} = \hat{x_i} / \sigma_{x_{i}}$,
where $\hat{x_i}$ is the average $x_i$ and $\sigma_{x_{i}}$ is the standard deviation. 

The overall nets are written into the hdf5 file. With large datasets and full augmentation we would expect there to be very little difference for the mean, $L^2$ and $\sigma$ for each element of the net. 

Where the test dataset is separate from the training dataset, or we have a generalisation dataset, the normalisation is done on the generalisation/test data using the map and values calculated from the training dataset.

\subsection{Spherical neural networks}

The spherical NN code is built in tensorflow 2,\cite{tensorflow} porting spherical convolutional layers from SpherePHD code\cite{lee2019spherephd} (as it's built in tensorflow 1). Essentially, the first convolutional layer is a standard tensorflow convolutional layer with a 3D input, with a connection table added so instead of being padded at the edges, the filters can wrap around the spherical input.

\subsection{Code and data}

The code to create the icospherical nets is on github under ICO: Icospherical Chemical Objects\cite{ICO}. It can take SMILES strings, mol2 and pdb files. The spherical nerual networks are available at github.com/ellagale/SphNN. The package comes with several different model types (regression, classification and autoencoder) with different numbers and types of inputs. The code is based on TensorFlow 2. 
Control experiments were done with DeepChem,\cite{wu2018moleculenet} and details on how to run these experiments are given in\cite{GaleTopolJournal} and DeepChem 2.3 validation\cite{DCval}. 

\subsection{Experimental methodology}

To test out the ICO as featurisations, the following experiments were performed.

\subsubsection{Experiment 1. Molecular properties}

\begin{figure}
    \centering
    \includegraphics[width=0.5\textwidth]{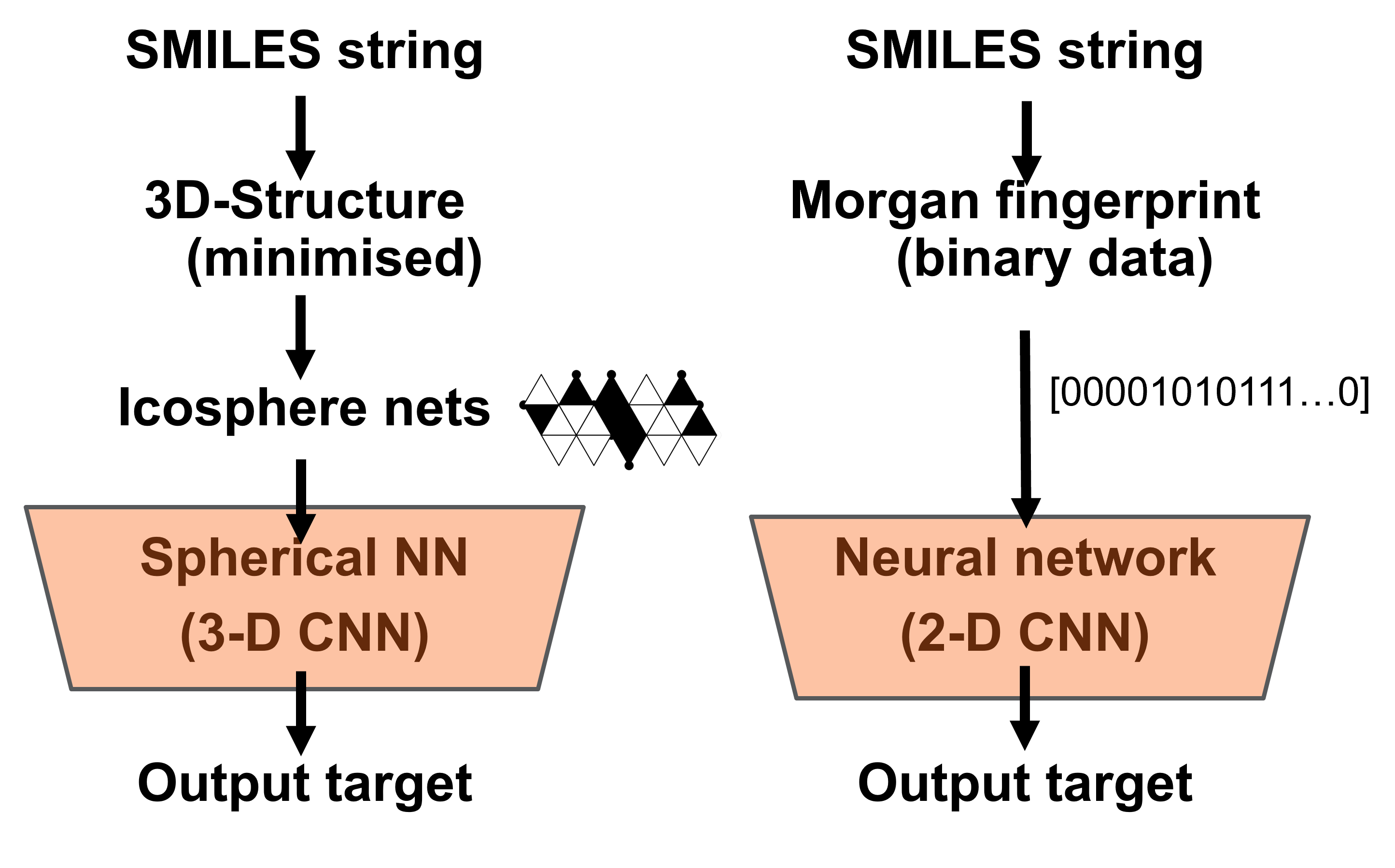}
    \caption{Architecture for experiment 1}
    \label{fig:prelim_arch}
\end{figure}

To demonstrate that this method works I undertook two preliminary experiments. To test that the SphNN could understand and match nets to the same molecule, SphNNs were trained on 80\% of the nets per molecule and tested their ability to identify the molecule from the input nets (10\% of the dataset). SphNN could recognise around 90\% of nets at the Ico-4 level and it was decided that this offered a good trade off between accuracy and size of dataset.


The original datasets were taken from molecular.ai
the molecular structure is loaded in as either a SMILES string or coordinates (mol file or pdb file). Rdkit is used to create (SMILES) or sanitize (pdb/mol) structures and located them in space and to calculate some simple features namely: number of atoms, number of bonds, number of non-hydrogens,  exact molar weight, octanol / water partition coefficient, number of heteroatoms, number of valence electrons, number of H acceptors, number of H donors,  principal moment of inertia 1 (smallest), principal moment of inertia 2, principal moment of inertia 3, spherocity, asphericity, eccentricity , inertial shape factor and radius of gyration, which are added to the output datafile. 

Final datafiles are output as hdf5 files and include the icospherical nets, rdkit calculated features and any information included from the original file (i.e. SMILES strings or other features). Files are indexed by molID\_ds which is a unique identifier for each molecule. 

This experiment did not use the analysis method described in ~\ref{ssec:analysis}.



\subsubsection{Experiment 2. Testing the effect of rotamers and conformers whilst doing solubility prediction using the Delaney dataset}

The data was taken from the MoleculeNet\cite{wu2018moleculenet} version of the Delaney dataset\cite{delaney2004esol}, which contains 1128 molecules.

The MoleculeNet version contained the following features: 
\begin{enumerate}
    \item ESOL predicted log solubility in mols per litre
    \item Minimum Degree
    \item Molecular Weight
    \item Number of H-Bond Donors
    \item Number of Rings
    \item Number of Rotatable Bonds	
    \item Polar Surface Area
    \item Measured log solubility in mols per litre	
    \item SMILES
\end{enumerate}

The ground truth $\hat{y}$ were the measured log solubility in mols per litre. SMILES string input was used to generate the nets as outlined earlier.


Two selection modes are implemented: ordered and random. Ordered takes the nets in order. The first 60 nets are rotations of the same conformer, so taking any number of nets less that 60 with ordered mode will give you only rotamers. The rotamers are in order, of if for example you take 10 nets, you will get 10 very similar but different in pixel space nets. Nets 60-120 are random rotations of different conformers (each nets was generated from a different instantiaion of the molecule from the smiles string, using RDKit . The random mode selects the nets randomly, giving you both rotamers and conformers, with a 50\% weighting towards the rotamers. Different numbers of nets were selected under both selection methods. The Delaney experiment used  the analysis method described in section~\ref{ssec:analysis}

\subsection{Experiment 3: Getting the most out of a small dataset whilst doing protein binding}

As protein binding can be considered a shape problem, it was decided to try the ICO featurisation on PDBBind. Level ICO-4 was used which can identify around 80\% of protein-ligands. With 20-30 epochs training. Obviously, a lot of information is compressed or lost from applying the ICO approach on such a big molecules as protein binding pockets. The usual approach is to train using the refined dataset (around 3000-4000 complexes) and then test on the core dataset (285 complexes). To test how well ICO does on small datasets, I switched it around and used the core dataset to train and the restricted dataset to test. PDBBind v2015 was used to compare to DeepChem 2.3 benchmarks.

\subsection{Analysis method\label{ssec:analysis}}

\begin{figure*}[htbp]
    \centering
    \includegraphics[width=0.5\textwidth]{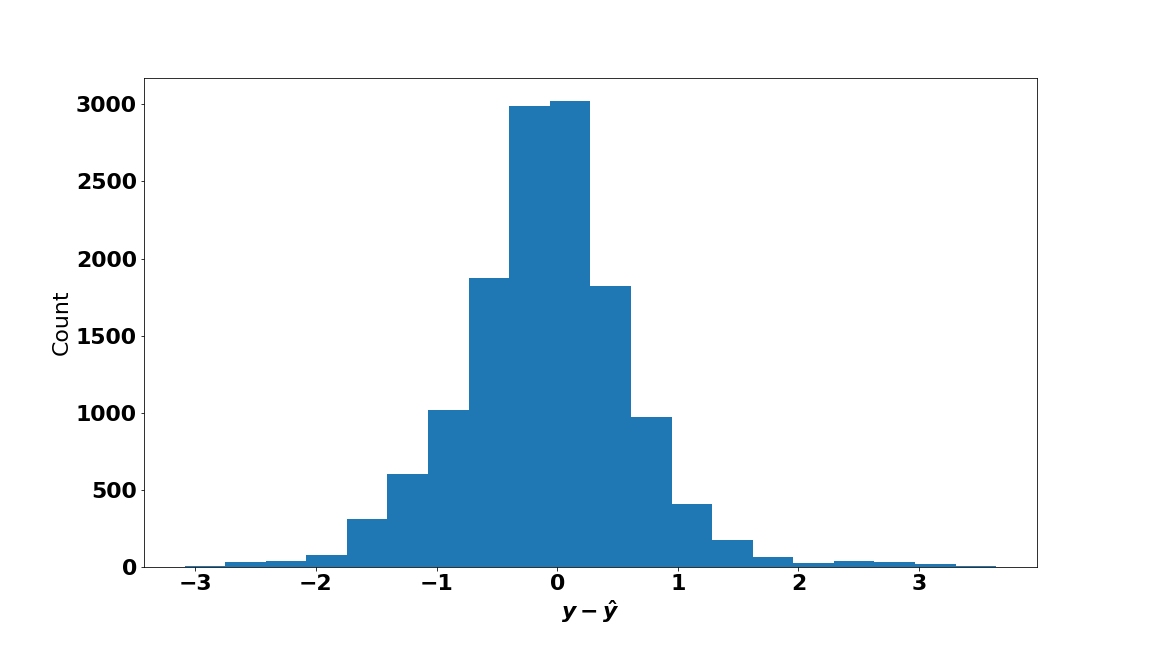} 
    \includegraphics[width=0.45\textwidth]{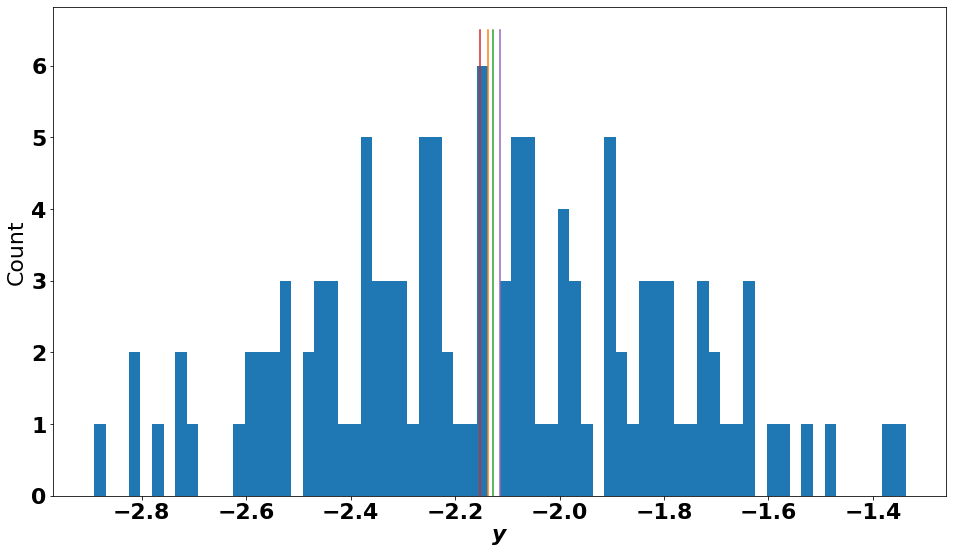}\\
    a. Errors for all points in this dataset b. Example  $\hat{y}$ data for a single $y$ in the data\\
    \includegraphics[width=0.45\textwidth]{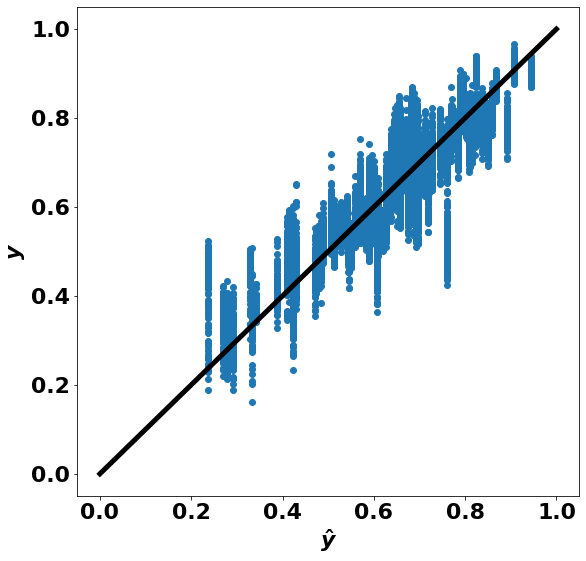} \includegraphics[width=0.5\textwidth]{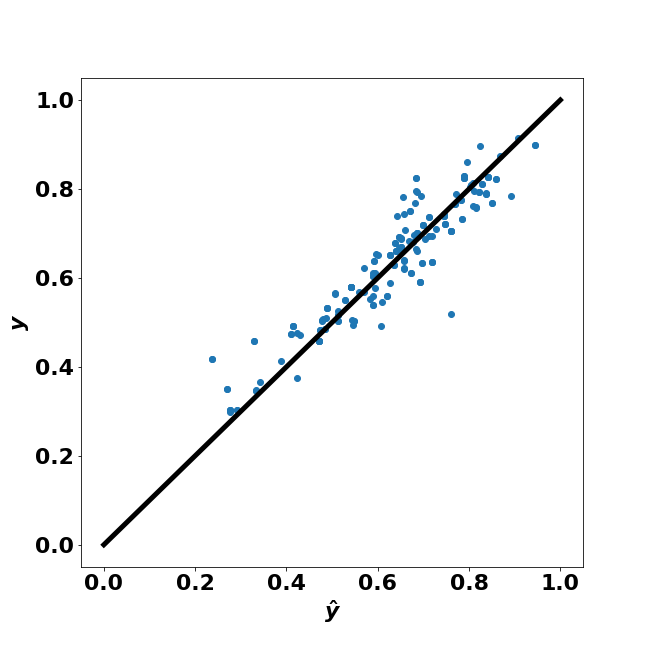}\\
    c. Normalised $y$-$\hat{y}$ data d.Cleaned $y$-$\hat{y}$ data
    \caption{The IQR ratio method of selecting predictions. a. errors between $y$ and $\hat{y}$ for a SphNN trained on the Delaney problem. The distribution is not normal, there are low peaks in the tails and symmetry is not assured. b. histogram of $\hat{y}$ values for a single datapoint. Orange and green lines: inner range, red purple: outer range. c. example normalised prediction data output from a SphNN. d. example normalised prediction data after the IQR data cleaning method.}
    \label{fig:analysis_method_2}
\end{figure*}

\begin{figure}[htp]
    \centering
    \includegraphics[width=0.45\textwidth]{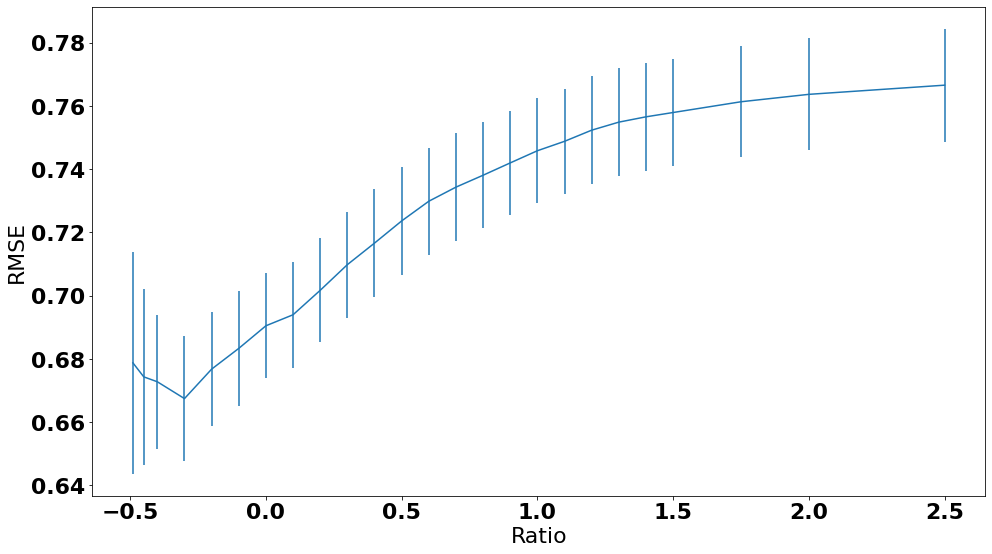} \\
    \includegraphics[width=0.45\textwidth]{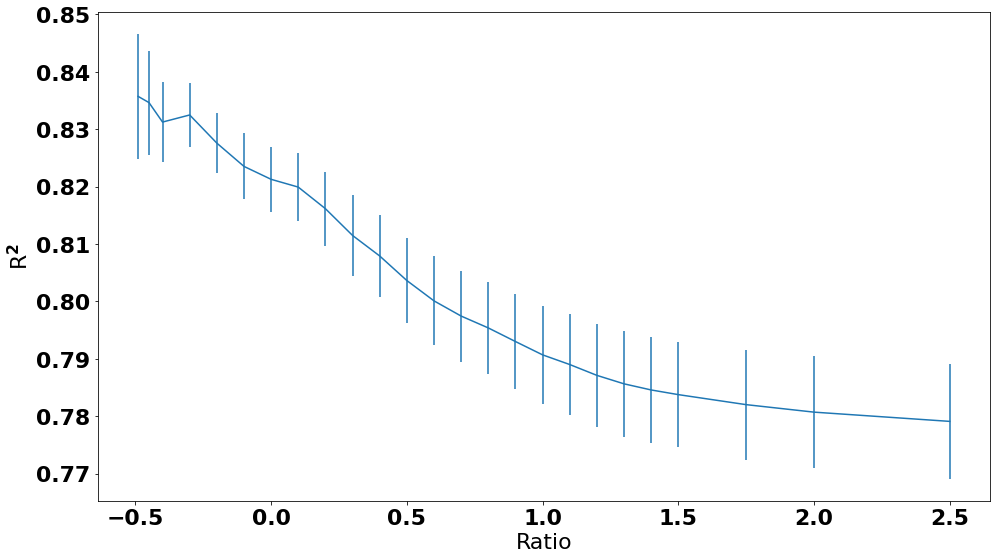}\\
    a. RMSE data against ratio b. R2 against ratio
    \caption{The effect of changing the ratio when cleaning the predicted data. Left: RMSE between $y$ and $\hat{y}$. Right: Pearson $R^2$ between $y$ and $\hat{y}$. Data taken from the run shown in figure~\ref{fig:norm_delaney}}
    \label{fig:analysis_method}
\end{figure}

There is a prediction for every input datapoint, and thus we get a spread of predictions, as shown in figure~\ref{fig:analysis_method_2}c. We can improve the output from the SphNNs by removing the outliers from this distribution, and this does not seem an unreasonable method, given how many NN researchers use fusion, average or majority voting with multiple models. The errors between $y$ and $\hat{y}$ are not normal, as shown in figure~\ref{fig:analysis_method_2}) and this has been confirmed via normality tests on the distribution. To remove outliers the following method is usually applied. Calculate a cut-off, $c$, from the interquartile range, $IQR$, and a ratio, $r$, as 
$$c = IQR \times r \:,$$

and calculate the lower and upper bounds from the first, $q_{25}$, and third, $q_{75}$,  quartile points using: 
$$
\mathrm{Lower bound} = q_{25} - c
$$
$$
\mathrm{Upper bound} = q_{75} + c \:.
$$ 
The usual value for $r$ is 1.5, We tested cutoffs from -0.49 to 2.5 (see figure~\ref{fig:analysis_method} and found that the best results were seen with $-0.49 \leq r < -0.1$. This method uses the IQR to determine the middle range of the distribution (which is more symmetrical) and then taking the rough midpoint of that distribution. (Fitting the curve to a distribution and finding the centre would most likely also work, but is more computationally intensive). Simply using the median value of the distributions (median averages are less sensitive to outliers than mean averages) does not improve the the results (see table~\ref{tab:cleaning}). Using extreme values of r (-0.49, -0.45) gives selects only a few datapoints and can chose values for which there are no datapoints, for example, see figure~\ref{fig:analysis_method_2}b where the range selected with $r=-0.49$ contains no datapoints. In this instance, we widen the range
and gives six datapoints to average. The cleaned predictions are show in figure~\ref{fig:analysis_method_2}d and table~\ref{tab:table}. In this work the value for $r$ of -0.45 was chosen based on this preliminary analysis.

\begin{table}[htbp]
    \centering
    \begin{tabular}{cp{1cm}p{1cm}p{1cm}}
    \hline
    Data analysis & $R^2$    & RMSE  & MAE   \\
    treatment   &    &   & \\
    \hline
    Original    & 0.805   & 0.711   & 0.527\\
    Median  & 0.779  (-3\%)   & 0.696 (-2\%) & 0.513 (-3\%) \\
    IQR method  & 0.875 (+9\%)    & 0.589 (-17\%) & 0.441 (-16\%)\\
    \hline
    \end{tabular}
    \caption{Output data cleaning improves the metrics. $r$ is -0.49.}
    \label{tab:cleaning}
\end{table}

\section{Results and Discussion}

\subsection{Experiment 1. Molecular properties}

Problems can be hard because the NN is underpowered or the dataset is too small
e.g.:
Classifying 1000 classes from 60,000 datapoints is harder than classifying 10 classes from 60,000 data points. In this section we go through some indicative results from the preliminary experiments, full results are in the S.I.




\begin{table}[htp]
    \centering
    \begin{tabular}{p{2cm}p{2cm}cp{1cm}}
    \hline
       Input  & Output & Augmentation & Accuracy on test \\
        
        \hline
        SMILES          & Charge  & No &  86.0\%  \\
        Icosphere level 2 &  Charge& No &  0.5\% \\
        Icosphere level 2& Charge  & Yes &   48.9\% \\
        \hline
    \end{tabular}
    \caption{Results from the preliminary classification experiment.}
    \label{tab:prelim_classification}
\end{table}

\begin{table*}[htp]
    \centering
    \begin{tabular}{|c|c|c|c|c|}
    \hline
       Input  & Output & Augmentation & MAE & $R^2$ \\
        $X$ & $y$ & & & \\
        \hline
        SMILES (control) & Exact molecular weight  & No &  124&  -18\\
        Icosphere level 2 & Exact molecular weight& No & 16.2&  0.83\\
        Icosphere level 2&Exact molecular weight  & Yes & 3.48  & 0.99\\
        \hline
        SMILES          & Radius of gyration & No & 1.1 &  -0.12\\
        Icosphere level 2 & Radius of gyration   & No & 0.258&  0.67\\
        Icosphere level 2& Radius of gyration  & Yes & 0.138 & 0.89  \\
        \hline
        SMILES          & MolLogP  & No & 1.39 &  -0.89\\
        Icosphere level 2 & MolLogP & No &0.816 & 0.461 \\
        Icosphere level 2&MolLogP   & Yes & 0.613 & 0.664 \\
                 \hline
    \end{tabular}
    \caption{Results from the preliminary regression experiment.}
    \label{tab:prelim_regression}
\end{table*}


We can now discuss the effects of choosing masses as the atom labels in the input. The radius of gyration (and other properties like moments of inertia) require the knowledge of both the masses and their position in space. As icospherical projection provides this information the calculation is easy, as evidenced by the high $R^2$ and low MAE on the task, with or without augmentation, see table~\ref{tab:prelim_regression}. Whereas the MACCS keys do not provide this information, and the control NN fails on the task. It might be possible for a very complex NN to be given enough MACCS keys for it to learn that C atoms have a mass, that carbons tend to arrange in specific ways and thus then learn how to calculate the radius of gyration, but this is an example of making the problem easier for the NN by giving it information in the correct domain to begin with. Compare this to the charge classification problem in table~\ref{tab:prelim_classification}. SMILES strings contain enough information to work out the charge and MACCS keys calculate it, so the information is given to the control NN and it suceeds easily. This demonstrates the importance of chosing the correct input featurisation for the problem, as if the charge on a molecule is important to the task, including it in the input greatly simplifies the task. However, notice how well augmentation helps the SphNN to solve this task as by going from the standard dataset to augmented the SphNN has gone from failing at the task (accuracy of near zero) to suceeding with about a 48\% accuracy. Another example: the ICO input includes the atomic weights, so to succeed at the exact molecular weight task, the SphNN only needs to learn how to count (summing the input of blue channel will give the result), and it's not surprising that it gets an $R^2$ of 0.99 on test. The SMILEs string input does not include that, and the MACCS keys does not count that, so it has a terrible $R^2$ on this task. Thus, when designing a system to solve a task, it is very important choose a featurisation method that contains the information necessary to solve the task in an accessible manner. Finally, the ICO method is able to do the MolLogP task, whereas the SMILES input cannot.

\subsection{Experiment 2: Solubility with the Delaney dataset}

\begin{figure}[htbp]
    \centering
    \includegraphics[width=0.4\textwidth]{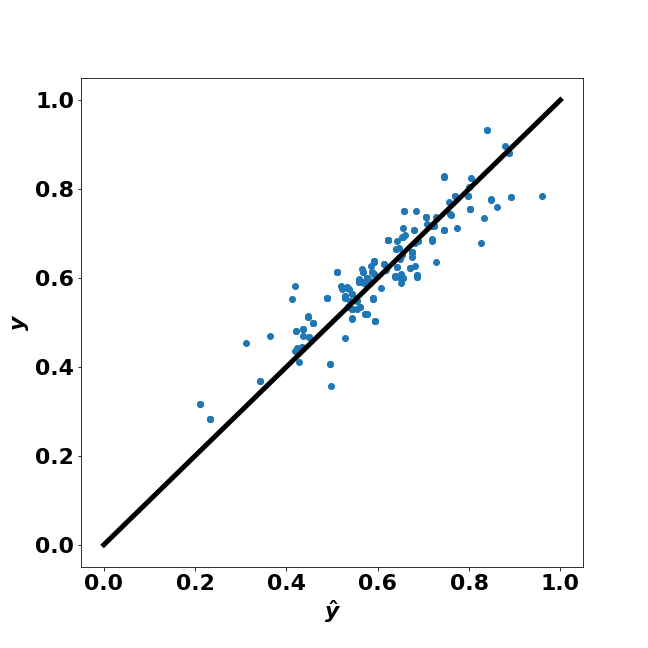}
    \caption{An example output from the Delaney experiment. Ground truth: $\hat{y}$, predicted output: $y$. Data normalised to be between 0 and 1. Cadence 120. Quick training.}
    \label{fig:norm_delaney}
\end{figure}

Results from a typical experiment are given in figure~\ref{fig:norm_delaney} and we can see that the SphNN does very well. 

\begin{figure}
    \centering
    \includegraphics[width=0.48\textwidth]{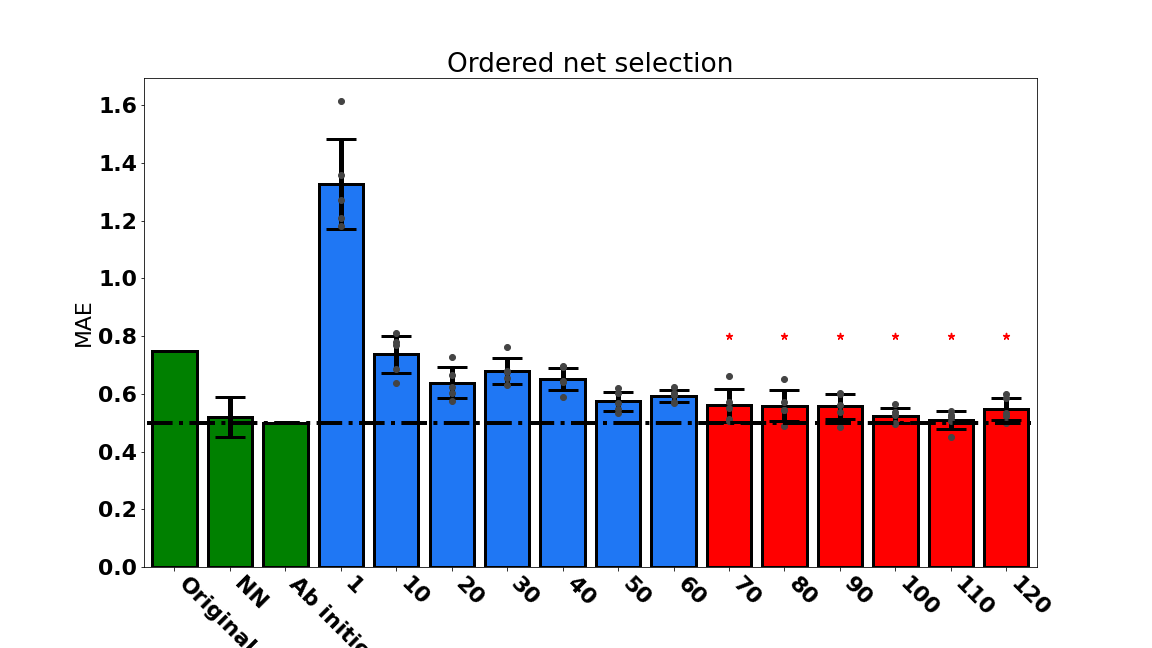}
    \includegraphics[width=0.48\textwidth]{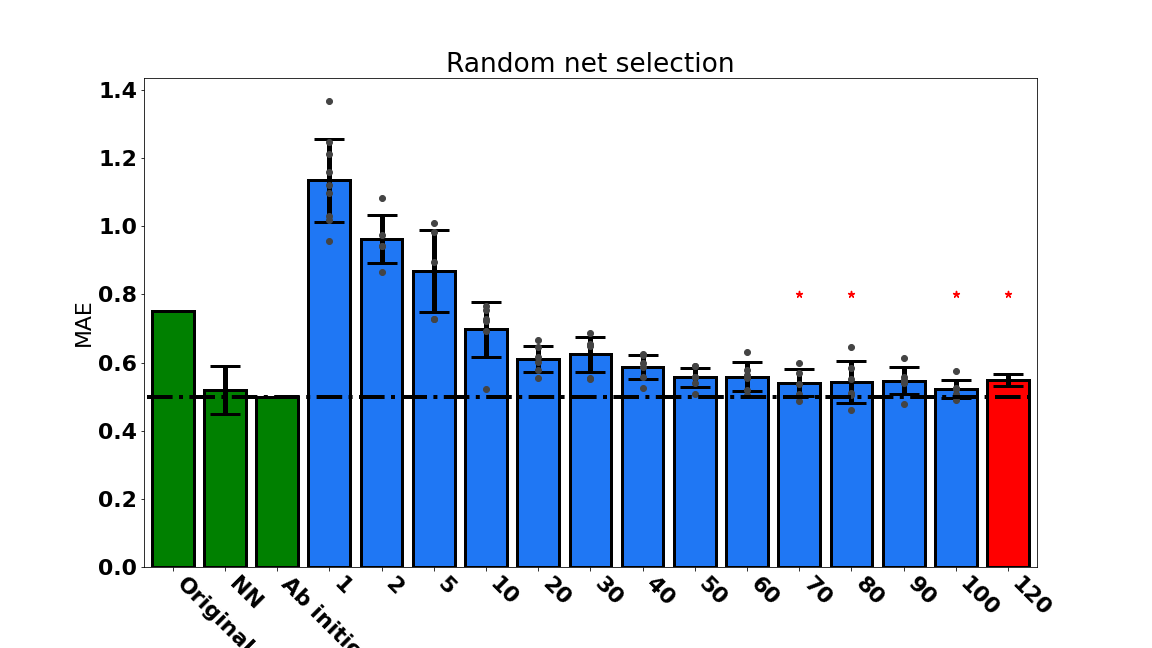}
    \caption{MAE for the Delaney dataset. Adding more nets and conformational information improves the results. Blue: rotations of a single conformer. Red: full set of rotations plus conformers at a random orientation. Data for more than 60 nets are statistically indistinguishable from the state of the art (QM). Original\cite{delaney2004esol}, NN\cite{duvenaud2015convolutional}, ab initio\cite{delaney2004esol}.}
    \label{fig:Delaney_MAE}
\end{figure}

The state of the art for the Delaney dataset is 0.5 and comes from QM methods, the average error in the experimental solubility data is around 0.6 log units.~\cite{jorgensen2002prediction} Figure~\ref{fig:Delaney_MAE} shows that ICO equal quantum mechanical methods, the control NN and experimental error on the dataset (i.e. we are at the limit of what is achievable with this dataset) and beat the Delaney equation to describe solubility. The ICO method works. Figure~\ref{fig:Delaney_MAE} shows that adding in conformers helps, as 60 rotations and only 10 conformers is enough too equal \textit{ab initio} results on the MAE. For the $R^2$ comparable results are acheived with only 5 or 10 nets. Generally, random selection is better, presumably as it covers a greater area of rotamer and conformer space.  These findings show that augmentation works. 

\begin{figure}
    \centering
    \includegraphics[width=0.48\textwidth]{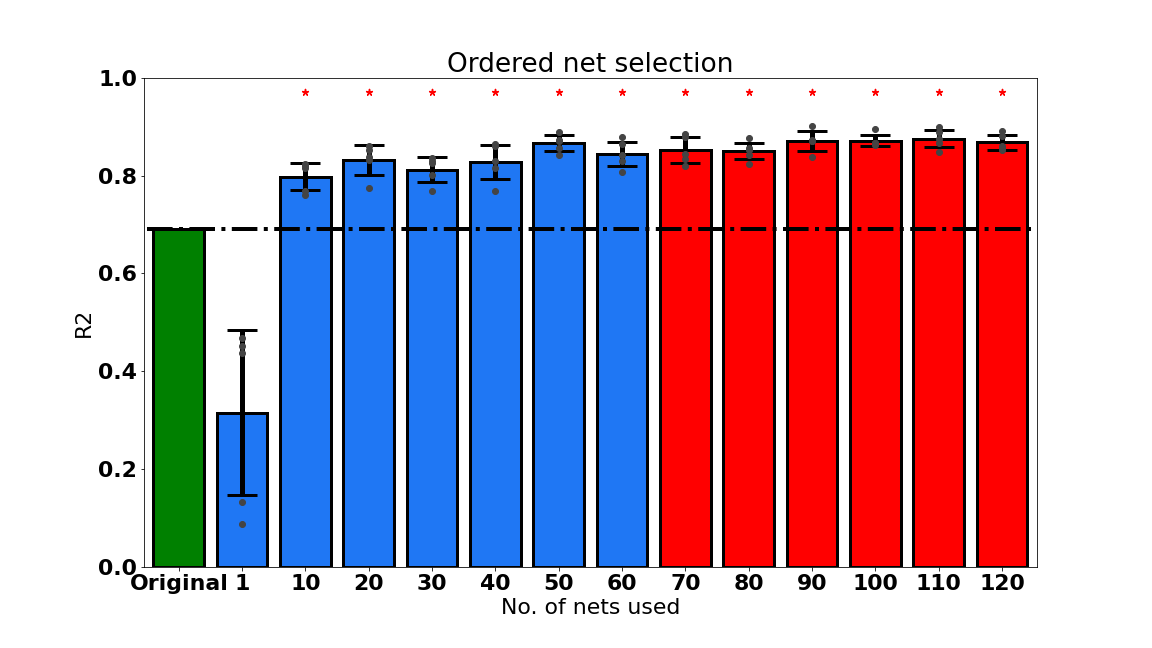}
    \includegraphics[width=0.48\textwidth]{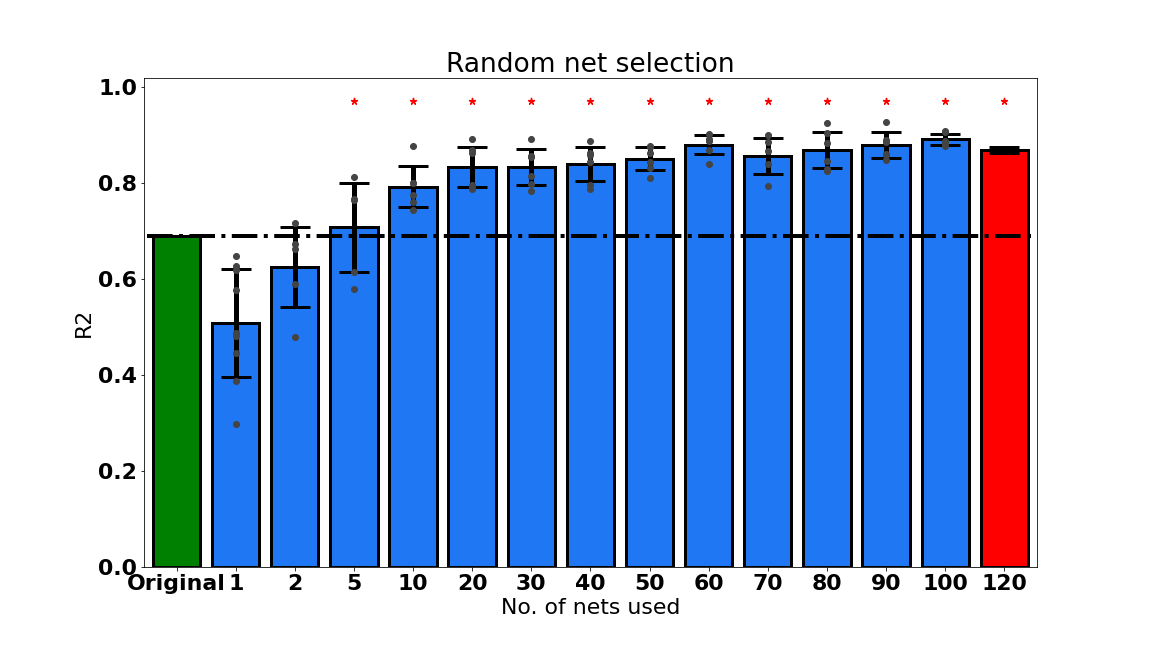}
    \caption{ICO's $R^2$ is much better than the state of the art. Blue: rotamers; Red: conformers}
    \label{fig:Delaney_R2}
\end{figure}

\subsection{Experiment 3: Getting the most out of small datasets using PDBBind}

\begin{figure}[htp]
    \centering
    \includegraphics[width=0.48\textwidth]{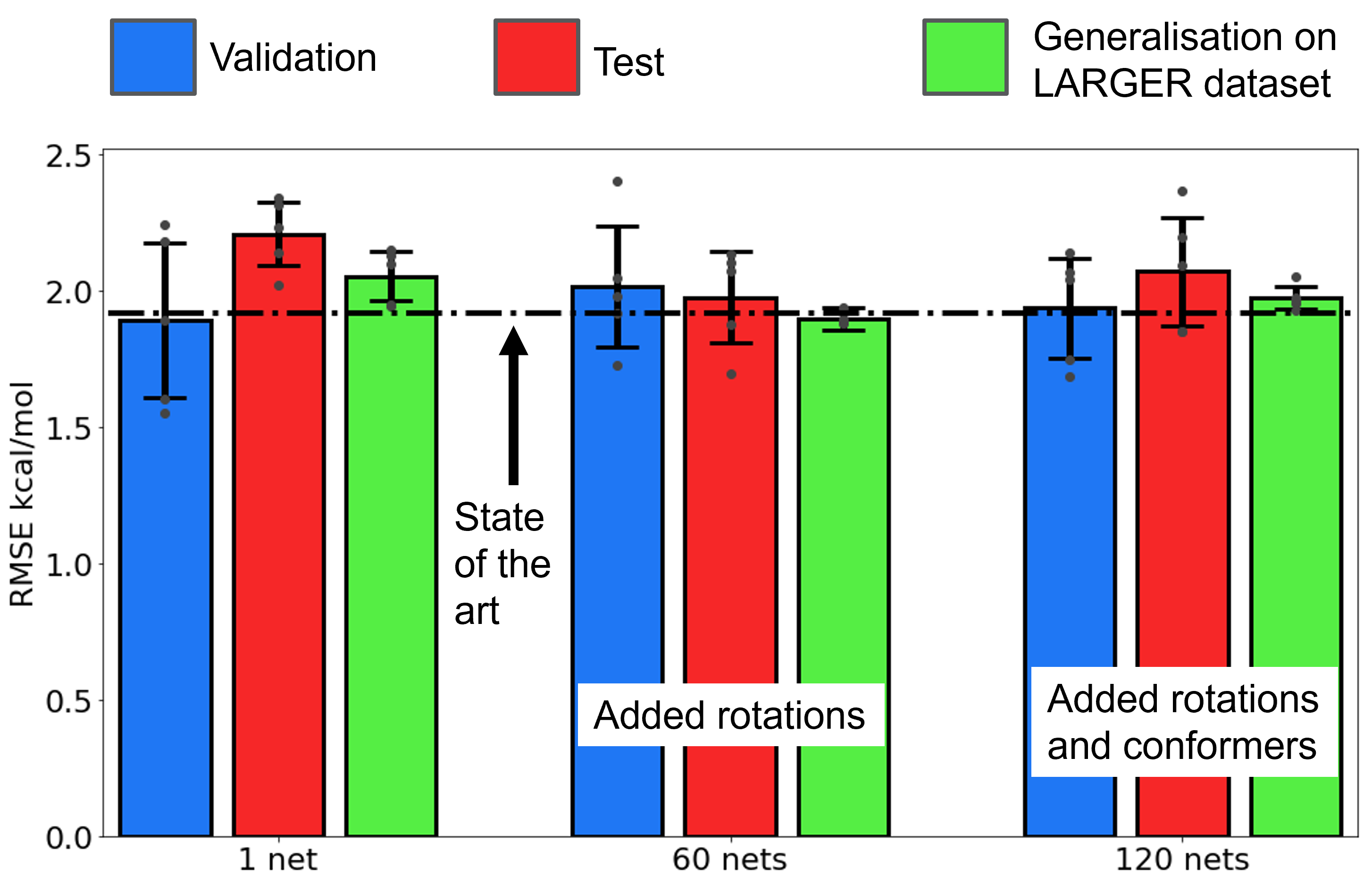}
    \caption{ICOSTAR gets the most out of small datasets. Task: training on PDBBind core and generalising the PDBBind restricted set (without core included).}
    \label{fig:train_core}
\end{figure}

Figure~\ref{fig:train_core} shows the results to using ICO and SphNN on PDBBind. With 1 net it is not too bad on test or generalisation, but we can see that adding augmentation from rotamers or rotamers and conformers brings it down to the state of the art (which is likely the limit on this dataset). The generalisation dataset is the restricted dataset, usually used as the training dataset, and we can see that the ICO augmentation technique has allowed us to get state of the art predictions on this dataset after training with only 285 complexes. 

\section{Conclusion}

In this paper I have presented a novel method of featurising molecular structure has the benefits of allowing us to use 3-D neural networks, retain rotational, translational and permutation invariance and offers a method for augmenting datasets. The ICO + SphNN method succeeds on the tasks. For example, we can predict solubility or MolLogP from just 3-D structure and atom masses. This method also offers the ability to augment small datasets and make it easier for the NN to learn what is important in the input data (the signal). In all cases, augmentation improved the predictions.  

In this work, I have argued that maintaining the rotational symmetry is better, but for these simple datasets, this method succeeds as well as the current state of the art, which is likely the limit for these well-studied datasets. Future work is looking for problems where the rotational symmetry is critical, and must be preserved, to investigate how this approach works. 

This work offers a 3-D featursation based only of 3-D structure and atom masses, that preserves chirality and rotational, translational and permutational symmetry. My other project\cite{GaleTopolJournal} offers a 3-D featurisation based only on 3-D structure that preserves topological invariants (gross shape features) and loses chirality. Work is already underway to try out the combination of both these 3-D features on shape-based problems, along with adding in some more `chemical' features like those in the rdkit featurisation to see if the combination works well on useful problems (like PDBBind or QM7). 

\section{Appendix}

\subsection{Mathematics\label{sec:maths}}

The interested reader is referred to `Geometric deep learning: Grids, groups, graphs, geodesics, and gauges'\cite{bronstein2021geometric} which aims to place neural networks on a coherent and geometrical foundation. The mathematics summarised in this section comes from that text.


A \textit{space} is a set of mathematical objects that can be treated as points (although they may not be points) and selected relationships between them (rules for that space). For example, the the positions of atoms may be treated as points in a space. The physical domain $\Omega$, space of signals $X(\Omega)$ and hypothesis class $F(X(\Omega))$ are all spaces.

A \textit{symmetry} of an object is a transformation that leaves a certain property of said object or system unchanged i.e. \textit{invariant}. Molecular properties are rotationally and translationally invariant. Symmetries can be invertible and the inverse is also a symmetry. hence the collection of all symmetries is a group.

\textit{Groups} are defined by how they \textit{compose} (i.e. can be stuck together). A \textit{group} is a set $\g{G}$ along with a binary operation called composition $\circ : \g{G} \times \g{G} \rightarrow \g{G}$ i.e. members of $\g{G}$ map into $\g{G}$, and satisfying the following axioms:

\begin{itemize}
    \item Associativity: $(\g{g} \circ \g{h}) \circ \g{l} = \g{g} \circ (\g{h} \circ \g{l}) \forall \g{h}$, where $ \g{h}, \g{l} \in \g{G}$.
    \item Identity: there exists a unique $\g{e} \in \g{G}$ satisfying $\g{e} \circ \g{g} = \g{g} \circ \g{e} = \g{g}$ $\forall \g{g} \in \g{G}.$
    \item Inverse: for each $\g{g} \in \g{G}$ there is a unique inverse $\g{g}^{-1} \in \g{G}$ such that $\g{g}\g{g}^{-1} = \g{g}^{-1}\g{g} = \g{e}$.
    \item Closure: the group is closed under composition i.e. for every $\g{g}, \, \g{h} \in \g{G}$ we also have $\g{g} \circ \g{h} \in \g{G}$.
\end{itemize}

Groups do not have to be commutative (Abelian groups are) so you can have $\g{g} \circ \g{h} \neq \g{h} \circ \g{g}$, and the group $SO(3)$ is noncommutative.


A $n$-dimensional real representation of the group $\g{G}$ is a \textit{map} $\rho : \g{G} \rightarrow \mathbb{R}^{n \times n}$ assigning to each $\g{g} \in \g{G}$ an invertible matrix $\rho(\g{g})$ which also satisfies $\rho(\g{g} \g{h}) = \rho(\g{g}) \rho(\g{h})\; \forall \g{g},\, \g{h} \in \g{G}$. 

Symmetries of the domain $\Omega$ are captured by group $\g{g}$ act on signals $x\in X(\Omega)$ through group representations $\rho(\g{g})$ and impose structure on the functions $f\in F(X(\Omega))$ which act on such signals. This is a powerful inductive bias reduces the space of possible interpolants $F(X(\Omega))$ to those which satisfy the symmetry priors. Thus, the $f \in F$ depend on both the signal $X(\Omega)$ and the domain $\Omega$.


\textit{Invariance} A function $f : X(\Omega) \rightarrow Y$ is $\g{G}$-invariant if $f(\rho(\g{g})x) = f(x),\: \forall \g{g} \in \g{G}$ and $x\in X(\Omega)$ i.e. its output is unaffected by the group action ($\rho$) on the input. For example, \textit{shift invariance}: translations of input do not matter, e.g. shift invariance in pooling layers in CNNs, doesn't matter where the object is in the frame. \textit{Rotational invariance} Rotation of the input does not matter, the physical orientation of a molecular structure is irrelevant for most chemical tasks. 

\textit{Equivariance:} A function $f: X(\Omega) \rightarrow X(\Omega)$ is $\g{G}$-equivariant if $f(\rho(\g{g})x) = \rho(\g{g}) f(x) \; \forall \g{g} \in \g{G}$, i.e. group action on the input affects the output in the same way. For example, \textit{Shift equivariance} A shift in the input causes an shift in the output by the same amount, used in pixel masks in image segmentation.





Space of moves on 2-D plane is isomorphic\footnote{Same shape, one can be related to the other.} to the plane. The space of rotations on a sphere are not isomorphic to a sphere, but the $SO(3)$ special orthogonal space in 3-D, which is all rotations of a sphere around the point at the centre and this is isomorphic to the real projective space, which is all the lines on a plane that pass through the origin. Thus, the spherical CNNs must work with both the sphere space $S^2$ and the $SO(3)$.

Convolutions require both the 2-D sphere $\Omega = S^2$ with the group of rotations $\g{G} = SO(3)$ special orthogonal group. We represent a point on the sphere as a 3D unit vector $||\vec{u}|| = 1$ the action of the group can be represented as $3\times3$ orthogonal matrix $\vec{R}$ with $det(\vec{R}) = 1$. The spherical convolution can then be written as  the inner product between the signal and rotated filter

$$
(x * \theta ) (\vec{R}) = \int_{\mathbb{S}^2} x(\vec{u}) \theta (\vec{R}^{-1} \vec{u}) d\vec{u} .
$$

The $SO(3)$ is a 3-D manifold and  preserves orientation of objects on the sphere. As the result of this convolution is not on the sphere, but $SO(3)$, so convolutions cannot be stacked and subsequent convolutions are defined on $SO(3)$ not $S^2$. 
 The next convolutional layer is then: \cite{bronstein2021geometric}

$$
((x*\theta)*\phi)(\vec{R}) = \int_{SO(3)} (x*\theta)(\vec{Q})\phi (\vec{R}^{-1} \vec{Q}) \mathrm{d}\vec{Q}.
$$







\section*{Acknowledgements}
E.G. would like to thank Taco S. Cohen for useful discussions, Oliver Matthews for proof-reading and EPSRC for funding.



\balance


\bibliography{references,bristol}
\bibliographystyle{rsc} 

\end{document}


\maketitle

\begin{abstract}

\end{abstract}

\section{Publicons for paper and supplementary information}

\section{Delaney Experiment}

measured log solubility in mols per litre

\subsection{$R^2$}

\begin{table}[]
    \centering
    \begin{tabular}{|c|c|}
        Name        & $R^2$ \\
        \hline
        Delaney Original\cite{delaney2004esol} & 0.69\\
         & \\
    \end{tabular}
    \caption{Caption .Units are log solubility in mols per litre.}
    \label{tab:Delaney_MAE}
\end{table}

\begin{table}[]
    \centering
    \begin{tabular}{c|l}
    no of nets & $R^2$ \\
    Delaney Original\cite{delaney2004esol} & 0.69\\
    1 & 0.32 $\pm$ 0.07\\
    10 &	0.80 $\pm$ 0.01\\
20 &	0.83 $\pm$ 0.014\\
30 &	0.81 $\pm$ 0.01\\
40 &	0.83 $\pm$ 0.01\\
50 &	0.87 $\pm$ 0.01\\
60 &	0.84 $\pm$ 0.01\\
70 &	0.85 $\pm$ 0.01\\
80 &	0.85 $\pm$ 0.001\\
90 &	0.87 $\pm$ 0.001\\
100 &	0.87 $\pm$ 0.005\\
110 &	0.88 $\pm$ 0.01\\
120 &	0.87 $\pm$ 0.006\\
    \end{tabular}
    \caption{Caption}
    \label{tab:my_label}
\end{table}

\begin{figure}
    \centering
    \includegraphics[width=\textwidth]{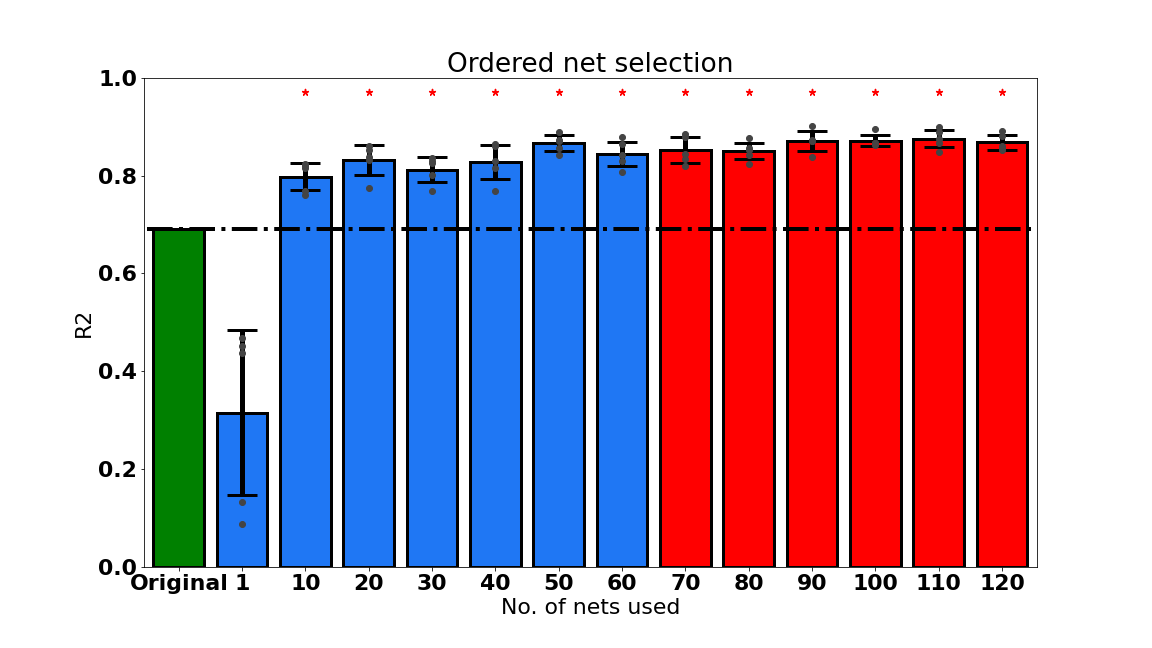}
    \includegraphics[width=\textwidth]{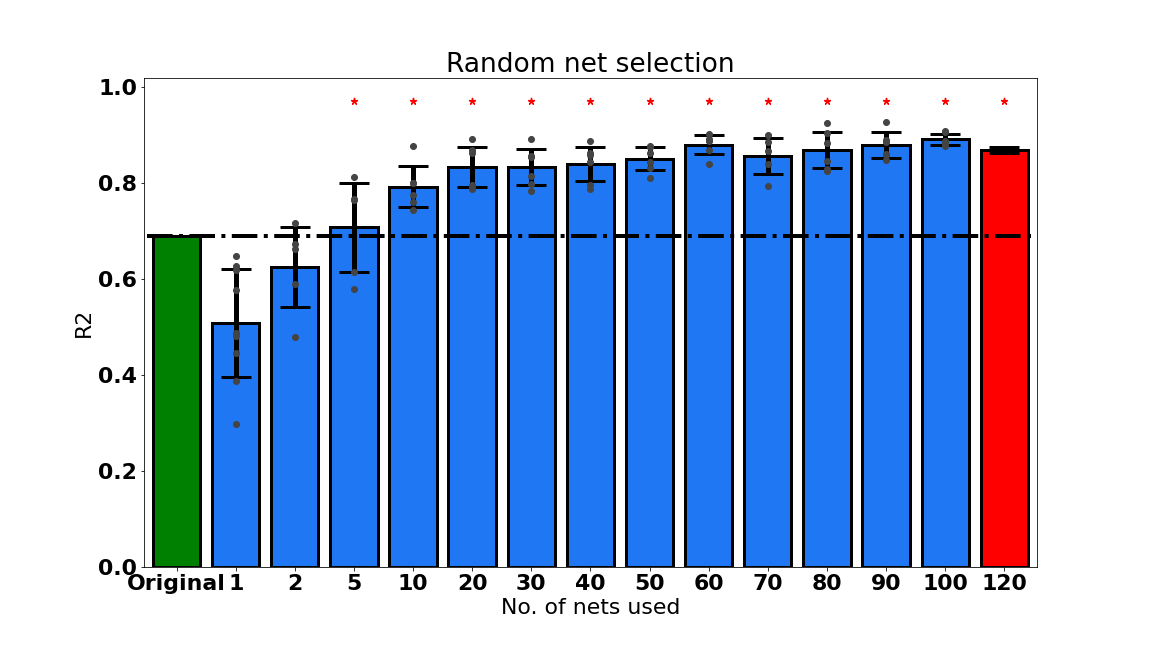}
    \caption{$R^2$ for the Delaney dataset. Original comes from the original paper\cite{delaney2004esol}.}
    \label{fig:Delaney_R2_ordered}
\end{figure}

\subsection{RMSE}

\begin{table}[]
    \centering
    \begin{tabular}{|c|c|}
        Name        & RMSE \\
        & log Mol/L\\
        \hline
        RF\cite{wu2018moleculenet} & $1.07 \pm 0.19$ \\
        Multitask\cite{wu2018moleculenet} & $1.12 \pm 0.15$\\
        XGBoost\cite{wu2018moleculenet} &  $0.99 \pm 0.14$\\
        KRR\cite{wu2018moleculenet} & $1.53 \pm 0.06$\\
        GC\cite{wu2018moleculenet} & $0.97 \pm 0.01$\\
        DAG\cite{wu2018moleculenet} & $0.82 \pm 0.08$\\
        Weave\cite{wu2018moleculenet} & $0.61 \pm 0.07$\\
        MPNN\cite{wu2018moleculenet} & $0.58 \pm 0.03$\\
         & \\
    \end{tabular}
    \caption{Results for the Delaney solubility dataset. Units are log solubility in mols per litre.}
    \label{tab:Delaney_R2}
\end{table}

 results for ordered (July) against DAG
40	0.15388267393536123	False same
50	0.05219976980978231	False
60	0.9674217548832718	False
70	0.34661235665577883	False
80	0.16178383044477837	False
90	0.055563846341909064	False
100	0.006613510810773841	True better
110	0.0030277145879711548	True
120	0.02730566231182435	True
the data is diffrent from Weave for the whole set, so not as good

\begin{figure}
    \centering
    \includegraphics[width=\textwidth]{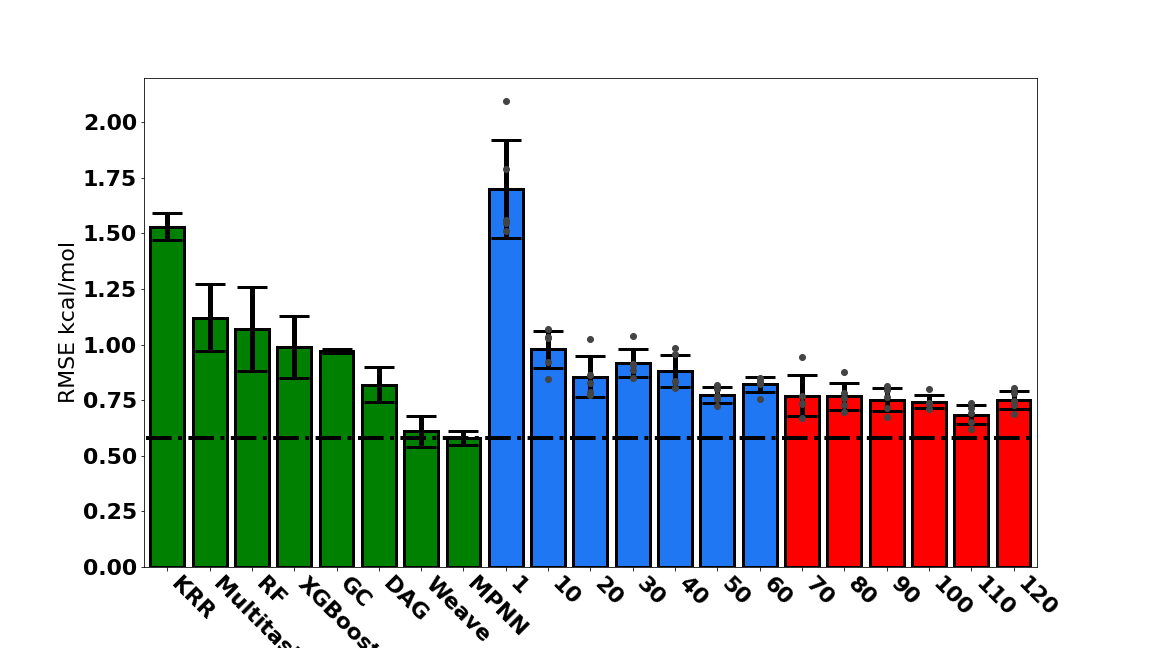}
    \includegraphics[width=\textwidth]{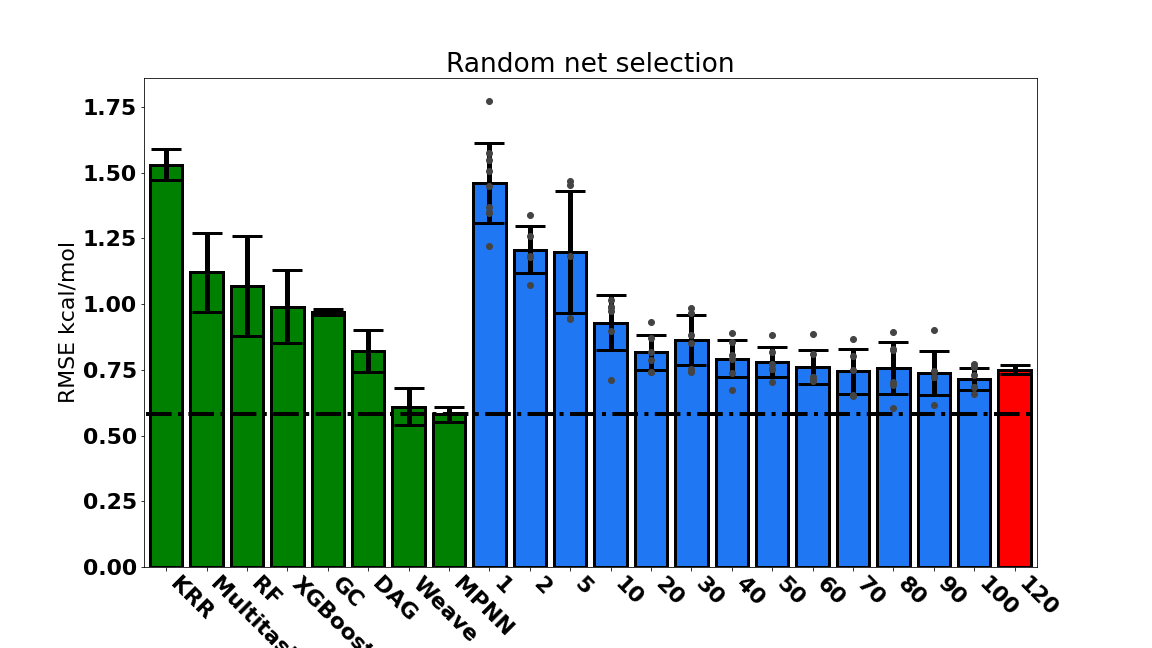}
    \caption{$R^2$ for the Delaney dataset. RF\cite{wu2018moleculenet},Multitask\cite{wu2018moleculenet},
        XGBoost\cite{wu2018moleculenet},
        KRR\cite{wu2018moleculenet},
        GC\cite{wu2018moleculenet},
        DAG\cite{wu2018moleculenet},
        Weave\cite{wu2018moleculenet},
        MPNN\cite{wu2018moleculenet}}
    \label{fig:Delaney_R2_ordered}
\end{figure}

\subsection{Delaney MAE}

REsults below are the raw results, they've not been cleaned! (As in, we've not used the thingy that removes the outliers and adds a couple of percent to the answers). 

\begin{table}[]
    \centering
    \begin{tabular}{|c|c|}
        Name        & MAE \\
        \hline
        Delaney Original\cite{delaney2004esol} & 0.75\\
        Neural FPs + neural net\cite{duvenaud2015convolutional}  & $0.52 \pm 0.07$\\
        \textit{Ab initio methods}\cite{delaney2004esol} & +/- 0.5 \\
    \end{tabular}
    \caption{Results for the Delaney solubility dataset. Units are log solubility in mols per litre. log Mol/L [[NTS may need proper ref for ab initio]]}
    \label{tab:Delaney_MAE}
\end{table}

\begin{figure}
    \centering
    \includegraphics[width=\textwidth]{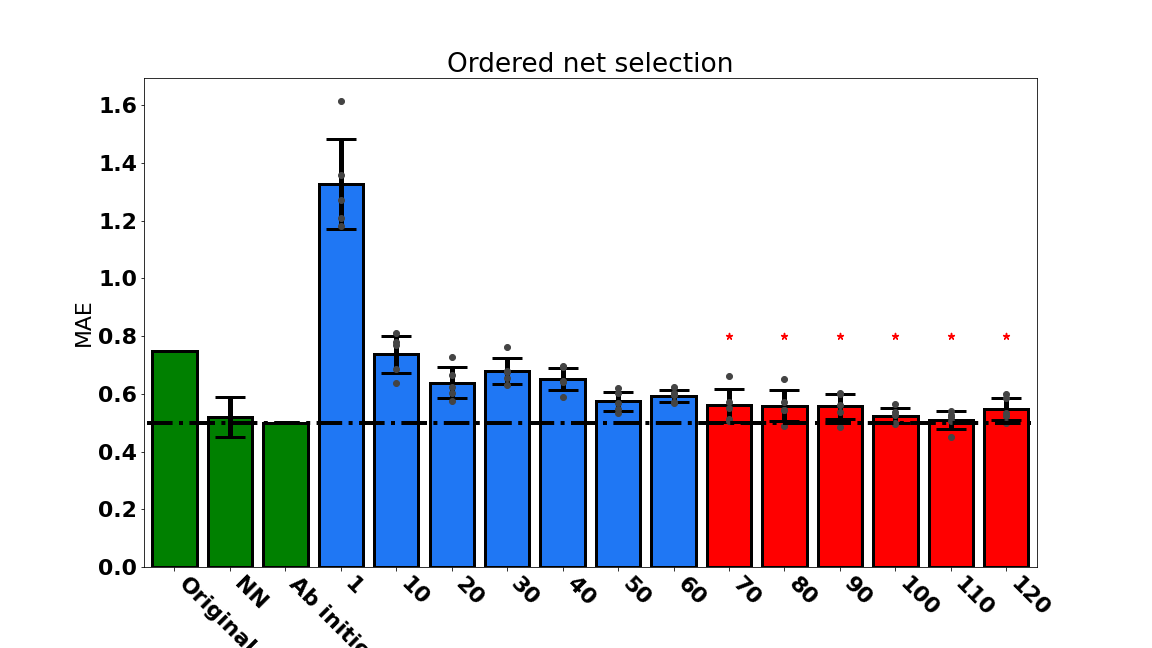}
    \includegraphics[width=\textwidth]{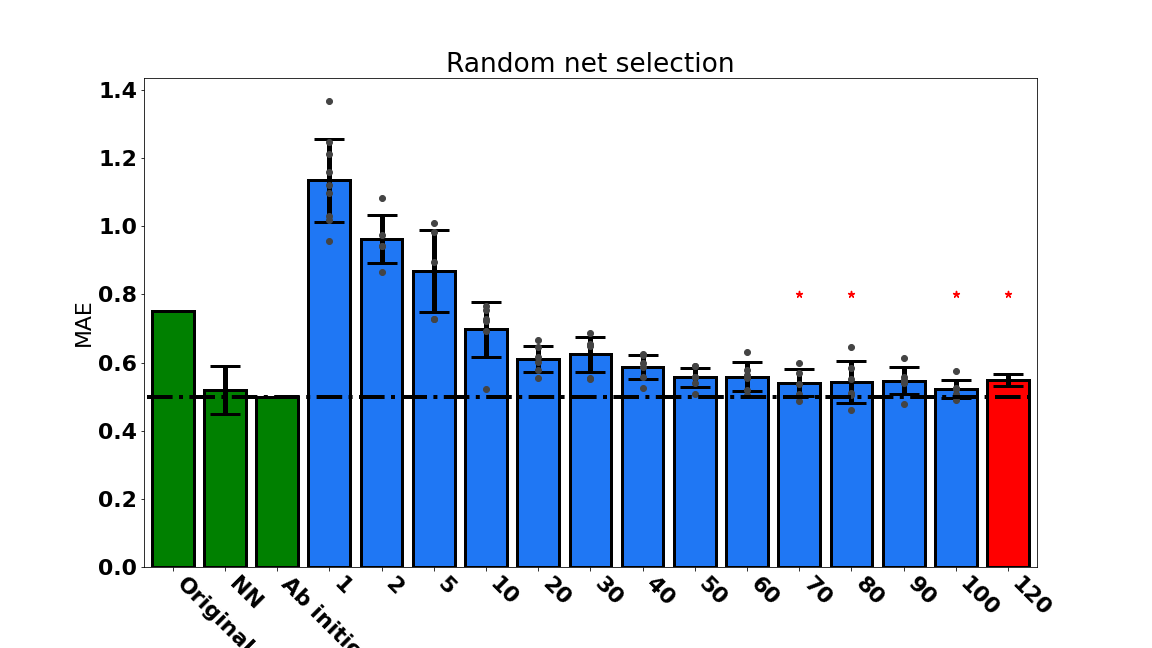}
    \caption{MAE for the Delaney dataset. Original\cite{delaney2004esol}, NN\cite{duvenaud2015convolutional}, ab initio\cite{delaney2004esol}.}
    \label{fig:Delaney_R2_ordered}
\end{figure}

do you want this from duvenaud paper? [Circular FPs + neural net $1.40 \pm 0.13$]

\subsection{Random order}

20	0.11685282910087347	False - as good as 120
30	0.08505334191671925	False
40	0.12250047174482624	False
50	0.1440049483227883	False
60	0.2792528821083871	False
70	0.5589782399549881	False
80	0.9843474994409236	False
90	0.460900856944095	False
100	0.007687791272077056	True - better tahn 120 (but only cos 120 is a bit lower)

\section{Experiment 1: Train and test on core dataset}

\subsection{Convolutional network only}

\begin{table}[]
    \centering
    \begin{tabular}{c|c|c|c|c|c|c|c|}
         & train r2 & val R2 & test R2 & gen r2 & test MAE & test MSE & test RMSE \\
         Random forest & 0.87$\pm0.005$  & 0.18$\pm 0.01 $ & 0.13$\pm$0.001 & -- & 1.46$\pm0.001$ & 4.03$\pm0.04$&2.0$\pm 0.01$ \\
         &  & & & & & &\\
         \hline
         &  & & & & & &\\
         &  & & & & & &\\

    \end{tabular}
    \caption{Experiment 1. Train on core dataset (80:10:10) split, generalise on restricted}
    \label{tab:Exp1}
\end{table}

\begin{table}[]
    \centering
    \begin{tabular}{|c|c|}
        name & R2 \\
        Random forest & \\
        NN? & \\
    \end{tabular}
    \caption{Pearson correlation coefficient for the PDBBind core dataset using a 80:10:10 train:validation:test split on a shuffled dataset, experiments were repeated 10 times (n=10), best value given in brackets. *score from the single best model [PDBBind v2016]}
    \label{tab:core_only}
\end{table}

\section{Experiment 4: Training on augmented and testing on core}

\begin{table}[]
    \centering
    \begin{tabular}{|c|c|c|c}
    \hline
        Name & trainable & no. & R$^2$ \\
            & parameters & desc.$^1$& \\
    \hline
    YAED-17f 3-head regression & ?& 17 & \textbf{0.92}\\
    yet another even deeper 3-head regression &  13mill & 15& \textbf{0.8997!}\\
    even deeper three-headed regression &  9,158,355  & 15 & \textbf{0.88}, 0.867$\pm ?$\\
    deep three headed regression & & 15 & 0.87 \\
    \textbf{deeper two-headed regression} & 181,425 & 0 & 0.85\\
    deep two-headed regression & 180,345 & 0 & 0.84\\
    two-headed regression model & & 0& ??\\
    \hline

    ES-IDM+ES-IEM GBT model\footnote{topological persistance features gradient based tree}\cite{meng2020persistent} & & & 0.840\\
    DeepAtom*\cite{rezaei2019improving} (augmented) &  & &  0.83\\
    K$_{deep}$*\cite{jimenez2018k} (augmented)[[cite orig paper]]  & 1,340,769 & & 0.82\\
    RF-Scorev3+RDKit ligand features$^{o}$\cite{boyles2020learning} & & & 0.821\\
    DeepAtom\cite{rezaei2019improving} &  & & 0.81\\
    RF-Score*\cite{rezaei2019improving} (augmented)[[cite orig paper]] & & & 0.81\\
    Mid-level Fusion model$^{+}$\cite{jones2020improved} & & 8& 0.810 \\
    ASFP \cite{zhang2020asfp} & & & 0.81 \\
    RF-Scorev3 features$^o$\cite{boyles2020learning} & & 185 & 0.814\\
    OnionNet \cite{zheng2019onionnet} & 143,882,777 & & 0.812($SD = 1.257$)\\
    RF-Score\cite{ballester2014does} & & & 0.803 \\
    3D-CNN (voxel) model$^{+}$\cite{jones2020improved} & & 8 or 21? & 0.723 \\
    Bappl+$^{\$}$\cite{soni2020improving} & & & 0.710\\
    DeepAtom\cite{rezaei2019improving} no hyperparameter refinement & & & 0.70\\
    SG-CNN model (graph)$^{+}$\cite{jones2020improved} & & 21 & 0.666 \\
    Best classical scoring function\cite{ballester2014does} X-Score::HMScore & & & 0.64 \\
    Mean classical scoring function\cite{ballester2014does} (n=18) & & & 0.427$\pm$0.03 \\
    Worst classical scoring function\cite{ballester2014does} SYBYL::F-Score & & & 0.22 \\
    cang2018representability & & & & 0.799\\
    \end{tabular}
    \caption{Pearson correlation coefficient (R$^2$) for test set using the standard protein binding training approach of train on the PDBBind refined dataset and test on the PDB core dataset and using with no validation. *score from the single best model.$^1$ Molecular descriptors refers to the number of molecular features input. In this count I do not count the coordinates, to include them increment the reported number for my work by 2 (protein coords and ligand coords). $^+$ this model was trained on both generalised and restricted dataset. $^{o}$ these models use the core dataset as both the valdiation and test sets. $^{\$}$ using RF to fit a MM forcefield [PDBBind v2016]}
    \label{tab:my_label}
\end{table}

\begin{table}[]
    \centering
    \begin{tabular}{|c|c|c|}
        reference & R2 & rmse \\
        li2022multiphysical; train 2016, test 2016\\
        TopBP   & 0.861 & \\
        li2022multiphysical 1D2DCNN & 0.799 & 1.91 \\
    \end{tabular}
    \caption{PCBBind 2016 results or train 2020 test 2016. }
    \label{tab:my_label}
\end{table}

\begin{table}[]
    \centering
    \begin{tabular}{|c|c|c|c}
    \hline
        Name &  R$^2$ \\
    \hline
    
    deeper two-headed regression & 0.81\\
    deep two-headed regression & \textbf{0.83}\\
    two-headed regression & ??\\    
    \hline
    DeepAtom*\cite{rezaei2019improving} (augmented)  & 0.79\\
    RF-Score*\cite{rezaei2019improving} (augmented)[[cite orig paper]]   & 0.75\\
    \end{tabular}
    \caption{Pearson correlation coefficient (R$^2$) for test set using the standard machine learning training approach of a randomly assigning complexes  80:10:10 train:validation:test datasets. The dataset used was a combination of the PDBBind refined and core datasets. *score from the single best model [PDBBind v2016]}
    \label{tab:my_label}
\end{table}

\begin{table}[]
    \centering
    \begin{tabular}{c|c|c|c}
    \hline
        Name & MAE \\
    \hline
    
    \hline 
    even deeper 3 headed regression & 0.539\\
    deep-3 headed regression & 0.588\\
    \hline
    OnionNet \cite{zheng2019onionnet} & 0.984 \\
     \cite{rezaei2019improving}  Autodock4[[rezaei2019improving:Morris2009]] & 2-3 kcal/mol\\
     ??[[rezaei2019improving:Gomes2017]]  & 1 kcal/mol\\
          Mid-level Fusion model$^{+}$\cite{jones2020improved} & 1.02 \\
     3D-CNN model$^{+}$\cite{jones2020improved} & 1.164 \\
     SG-CNN model$^{+}$\cite{jones2020improved} & 1.321 \\
     AK-score single\cite{kwon2020ak} & 1.511\\
     \hline
     AK-score ensemble\cite{kwon2020ak} & 1.101 \\ 
    \end{tabular}
    \caption{Caption}
    \label{tab:my_label}
\end{table}

\begin{table}[]
    \centering
    \begin{tabular}{|c|c|c|c}
    \hline
        Name & RMSE \\
    \hline
        even deeper 3 headed regression & 0.777\\
    deep-3 headed regression & 0.795\\
    \hline
    OnionNet \cite{zheng2019onionnet} & 1.278 \\
     DeepAtom*\cite{rezaei2019improving} (augmented)   & 1.1\\
     RF-Score*\cite{rezaei2019improving} (augmented)[[cite orig paper]]  & 1.2\\
     K$_{deep}$*\cite{ jimenez2018k} (augmented)[[cite orig paper]]  &  1.27\\
          Mid-level Fusion model$^{+}$\cite{jones2020improved} & 1.308 \\
     3D-CNN model$^{+}$\cite{jones2020improved} & 1.501 \\
     SG-CNN model$^{+}$\cite{jones2020improved} & 1.650 \\
     AK-score single\cite{kwon2020ak} & 1.415\\
     \hline
     AK-score ensemble\cite{kwon2020ak} & 1.293 \\ 
     
    \end{tabular}
    \caption{RMSE for test set using the standard protein binding training approach of train on the PDBBind refined dataset and test on the PDB core dataset and using with no validation. *score from the single best model [PDBBind v2016]}
    \label{tab:my_label}
\end{table}

\begin{table}[]
    \centering
    \begin{tabular}{|c|c|c|c|c|c|c|c|c|}
         train 1 & split 1 & shuffle & train 2 & split 2 & shuffle & gen 1 & gen 2 & epochs needed  \\
         \hline
         100\% & 80:10:10 & all & n/a & n/a & n/a& n/a & 0.915 & 29\\
         100\% & 80:10:10 & molecule & n/a & n/a & n/a & n/a &  & \\ 
        100\% & 80:10:10 & all & 100\% & 80:10:10 & molecule & n/a &  & \\ 
         75\% & 80:10:10 & all & 80:10:10 & 100\% & all & * &  0.904 & 64 (36+28)\\
        75\% & 80:10:10 & all & 80:10:10 & 100\% & molecule & * &  0.880 & 44 (32+12)\\
        50\% & 80:10:10 & all & 80:10:10 & 50\% & molecule & * &   & \\

    \end{tabular}
    \caption{The effect of training method and shuffling mode. *Needs to be gen tested.}
    \label{tab:training_method}
\end{table}

\clearpage

\section{Methods}

\begin{table}[]
    \centering
    \begin{tabular}{c|ccc}
        Network name & no. of layers & layer types & no of trainable parameters\\
        \hline
        sp-3 & 3 layers & sp-conv, average & \\
        deep two in net & 7 & sp-conv, fc, max pool, flatten, concat &\\
    \end{tabular}
    \caption{Layer count does not include flatten, concatiation or average pooling.}
    \label{tab:my_label}
\end{table}

\bibliographystyle{plain}
\bibliography{references}